%% file: main.tex
\documentclass{article}
\IfFileExists{iclr2027_conference_OFF.sty}{\usepackage{iclr2027_conference}\usepackage{times}\usepackage{natbib}}{\IfFileExists{iclr2026_conference_OFF.sty}{\usepackage{iclr2026_conference}\usepackage{times}\usepackage{natbib}}{\IfFileExists{neurips_2026_OFF.sty}{\usepackage[preprint]{neurips_2026}}{\usepackage[numbers]{natbib}\usepackage[margin=1in]{geometry}}}}
\usepackage{caption}

\usepackage{algorithm}
\usepackage{algorithmic}
\usepackage{amsmath,amssymb,amsfonts,amsthm,mathtools}
\usepackage{graphicx}
\usepackage{booktabs}
\usepackage{microtype}
\usepackage{subcaption}
\usepackage{adjustbox}
\usepackage{tikz}
\usepackage{xcolor}
\usepackage{placeins}
\usepackage{float}
\usepackage{hyperref}

\usetikzlibrary{arrows.meta,positioning,calc,fit,backgrounds,shapes.geometric}

\usepackage{tcolorbox}
\usepackage{pgfplots}\pgfplotsset{compat=1.18}

\newcommand{\E}{\mathbb{E}}
\newcommand{\R}{\mathbb{R}}
\newcommand{\Var}{\mathrm{Var}}
\newcommand{\KL}{\mathrm{KL}}

\newtheorem{theorem}{Theorem}
\newtheorem{lemma}{Lemma}
\newtheorem{proposition}{Proposition}
\newtheorem{corollary}{Corollary}
\theoremstyle{definition}

\theoremstyle{remark}
\newtheorem{remark}{Remark}

\definecolor{cblue}{RGB}{59,130,246}
\definecolor{cgreen}{RGB}{34,197,94}
\definecolor{corange}{RGB}{249,115,22}
\definecolor{cred}{RGB}{239,68,68}
\definecolor{cpurple}{RGB}{168,85,247}
\definecolor{cteal}{RGB}{20,184,166}

\newcommand{\RawExp}{\textsc{wrapped-Exp}}

\newcommand{\bExp}{\mathrm{bExp}}

\newcommand{\vol}{\mathrm{vol}}
\usepackage{enumitem}
\newcommand{\RR}{\mathbb{R}}
\newcommand{\SSS}{\mathbb{S}}
\newcommand{\HH}{\mathbb{H}}
\newcommand{\Exp}{\operatorname{Exp}}

\title{Radial Compensation: The Inverse Base-Distribution Problem for Chart-Based Generative Models on Riemannian Manifolds}
\author{Marios Papamichalis\thanks{\raggedright Human Nature Lab, Yale University, New Haven, CT 06511, USA. \href{mailto:marios.papamichalis@yale.edu}{\nolinkurl{marios.papamichalis@yale.edu}}}
\and Regina Ruane\thanks{\raggedright Department of Statistics and Data Science, The Wharton School, University of Pennsylvania, Philadelphia, PA, USA. \href{mailto:ruanej@wharton.upenn.edu}{\nolinkurl{ruanej@wharton.upenn.edu}}}}

\date{}

\usepackage{tikz}
\usepackage{xcolor}
\usepackage{graphicx}

\usetikzlibrary{
  arrows.meta,
  calc,
  positioning
}

\usepackage{tikz}
\usepackage{xcolor}
\usepackage{graphicx}
\usetikzlibrary{arrows.meta,positioning,calc,fit,backgrounds}

\usepackage{tikz}
\usepackage{xcolor}
\usepackage{caption}
\usetikzlibrary{arrows.meta,positioning,calc}

\newcommand{\ars}{\iota}  

\definecolor{cblue}{RGB}{59,130,246}
\definecolor{cgreen}{RGB}{34,197,94}
\definecolor{corange}{RGB}{249,115,22}
\definecolor{cred}{RGB}{239,68,68}
\definecolor{cpurple}{RGB}{168,85,247}
\definecolor{cteal}{RGB}{20,184,166}

\begin{document}

\maketitle
\begin{abstract}
Latent-variable models on spheres and hyperbolic spaces usually draw a Gaussian in the tangent space at a base point and push it onto the manifold. On these spaces the distance from the base point is the coordinate that carries meaning: depth in a hierarchy, the angle of a rotation, the deviation of a protein frame from a reference. We show that the standard construction silently replaces whatever distance distribution the modeler intended with a fixed one, a scaled chi law whose shape no setting of the scale can change. We then solve the reverse problem. Given the intended distance distribution, we derive in closed form the tangent density that realizes it, prove it is the only isotropic choice with chart-independent likelihoods for a broad class of charts, and prove a lower bound with explicit constants on what ignoring the problem costs a variational autoencoder. Experiments backed by an exact per-run normalization audit confirm that the compensated prior is invariant to the chart and stable across scales, every wrapped baseline we train collapses to the boundary of its chart, curvature becomes recoverable where the wrapped prior fails and protein-orientation likelihood improves from 2.58 to 0.87 nats at identical accuracy.
\end{abstract}

\section{Introduction}
\label{sec:intro}

Latent-variable generative models on curved spaces are usually built by wrapping \citep{nagano2019wrapped,bose2020latent,skopek2019mixed,galaz2022wrapped}: draw a vector in the tangent space at a pole $p$, typically from a Gaussian, and map it to the manifold with the exponential map. The construction is popular because it reuses Euclidean densities, reparameterization, and flow architectures. It has a semantic defect. On a sphere or a hyperbolic space, the coordinate that carries meaning is the geodesic radius $R=d(p,q)$: the depth of a concept in a hyperbolic hierarchy, the angle of a rotation on $\SSS^3$, the distance of a protein frame from a reference. A modeler who sets the tangent scale $\sigma$ believes they are choosing the distribution of $R$. They are not. The exponential map preserves radii but not volume, so the realized law of $R$ is the scaled chi law $\sigma\chi_n$: its shape is fixed by the latent dimension $n$, its mean grows as $\sigma\sqrt n$, and its relative spread falls as $n^{-1/2}$ (below $9\%$ at $n=64$). Whatever radial structure was intended, the model concentrates its mass in a thin shell (Figure~\ref{fig:rc_pipeline_clean} gives the distribution-level picture).

\begin{figure*}[t]
\centering
\scalebox{0.72}{\begin{tikzpicture}[
  x=1cm,y=1cm,
  >=Latex,
  font=\small,
  panel/.style={draw, rounded corners=2pt, line width=0.55pt, fill=none},
  title/.style={font=\bfseries\small, align=left, text width=4.45cm, anchor=north west},
  ptxt/.style={font=\scriptsize, align=center, text width=4.30cm},
  arrlbl/.style={font=\scriptsize, fill=white, inner sep=1.5pt, align=center}
]

\colorlet{rawc}{blue!70}
\colorlet{rcc}{green!55!black}
\colorlet{geomc}{black!75}

\draw[panel] (0.00,0.00) rectangle (5.10,5.35);
\draw[panel] (5.55,0.00) rectangle (10.65,5.35);
\draw[panel] (11.10,0.00) rectangle (16.20,5.35);

\node[title] at (0.18,5.08) {1. Choose the target\\geodesic-radius law};
\node[title] at (5.73,5.08) {2. RC prewarps the\\tangent-space base};
\node[title] at (11.28,5.08) {3. Map through any\\azimuthal chart};

\draw[->, line width=0.45pt] (0.72,1.35) -- (3.08,1.35);
\draw[->, line width=0.45pt] (0.72,1.35) -- (0.72,3.15);

\draw[rcc, line width=0.90pt, smooth]
  plot coordinates {
    (0.80,1.40)
    (1.00,1.48)
    (1.25,1.78)
    (1.48,2.30)
    (1.72,2.75)
    (1.96,2.98)
    (2.20,2.72)
    (2.44,2.18)
    (2.68,1.70)
    (2.98,1.40)
  };

\node[font=\scriptsize\itshape, rcc] at (1.92,3.42) {$R \sim p_R$};
\node[font=\scriptsize] at (1.92,1.02) {$R=d(p,q)$};

\node[ptxt] at (2.55,0.46)
{distance from the pole in geodesic units\\
(e.g.\ HalfNormal, Gamma, TruncNormal)};

\begin{scope}[shift={(8.10,3.05)}]
  \draw[->, line width=0.45pt] (-1.10,0) -- (1.15,0);
  \draw[->, line width=0.45pt] (0,-1.10) -- (0,1.15);

  \draw[dashed, rawc, line width=0.70pt] (0,0) circle (0.56);

  \draw[rcc, line width=0.95pt] (0,0) circle (0.88);

  \node[font=\scriptsize] at (1.18,0.98) {$T_p\mathcal M$};
\end{scope}

\draw[->, line width=0.40pt] (6.15,0.98) -- (7.95,0.98);
\draw[->, line width=0.40pt] (6.15,0.98) -- (6.15,1.88);

\draw[dashed, rawc, line width=0.60pt, smooth]
  plot coordinates {
    (6.24,1.04)
    (6.42,1.25)
    (6.60,1.47)
    (6.82,1.68)
    (7.02,1.60)
    (7.20,1.35)
    (7.36,1.08)
  };

\draw[rcc, line width=0.82pt, smooth]
  plot coordinates {
    (6.58,1.02)
    (6.82,1.12)
    (7.08,1.40)
    (7.34,1.66)
    (7.60,1.56)
    (7.82,1.24)
    (7.94,1.04)
  };

\node[font=\scriptsize, rawc] at (6.82,2.05) {raw base};
\node[font=\scriptsize, rcc]  at (7.58,1.82) {RC base};

\node[ptxt] at (8.10,0.44)
{raw base $\neq$ RC base\\
RC changes the tangent distribution\\
so the chosen law of $R$ is recovered};

\begin{scope}[shift={(14.00,3.00)}]
  \draw[->, line width=0.45pt] (-1.55,0.45) -- (-0.92,0.45);
  \draw[->, line width=0.45pt] (-1.55,0.05) -- (-0.92,0.05);
  \draw[->, line width=0.45pt] (-1.55,-0.35) -- (-0.92,-0.35);

  \node[font=\scriptsize, anchor=east] at (-1.62,0.45) {Exp};
  \node[font=\scriptsize, anchor=east] at (-1.62,0.05) {$\mathrm{bExp}_{\alpha}$};
  \node[font=\scriptsize, anchor=east] at (-1.62,-0.35) {GCL};

  \draw[geomc, line width=0.70pt] (0.30,0) circle (0.90);

  \fill[geomc] (0.30,0.90) circle (0.03);
  \node[font=\scriptsize, above] at (0.30,0.90) {$p$};

  \fill[rcc!12] (0.30,0.22) circle (0.35);
  \draw[dashed, rcc, line width=0.55pt] (0.30,0.22) circle (0.35);

  \draw[densely dashed, geomc, line width=0.45pt] (0.30,0.90) -- (0.30,0.57);

  \node[font=\scriptsize\itshape, rcc] at (0.30,-0.52) {$R \sim p_R$};
\end{scope}

\node[ptxt] at (13.65,0.44)
{same chart family\\
same law of $R$ on $\mathcal M$\\
conditioning only};

\draw[->, line width=0.70pt] (5.10,2.70) -- (5.55,2.70);
\node[arrlbl] at (5.325,3.12) {build RC\\base};

\draw[->, line width=0.70pt] (10.65,2.70) -- (11.10,2.70);
\node[arrlbl] at (10.875,3.12) {choose\\chart};

\draw[dashed, rawc, line width=0.70pt] (1.20,-0.18) -- (1.85,-0.18);
\node[font=\scriptsize, anchor=west] at (1.95,-0.18) {raw / uncompensated};

\draw[rcc, line width=0.90pt] (5.65,-0.18) -- (6.30,-0.18);
\node[font=\scriptsize, anchor=west] at (6.40,-0.18) {target / RC-correct};

\draw[geomc, line width=0.70pt] (10.80,-0.18) -- (11.45,-0.18);
\node[font=\scriptsize, anchor=west] at (11.55,-0.18) {chart / manifold geometry};

\draw[rounded corners=2pt, line width=0.45pt]
  (0.95,-1.25) rectangle (15.25,-0.38);

\node[font=\scriptsize, align=center, text width=13.3cm] at (8.10,-0.82) {RC fixes the law of distance on the manifold first; after that, the chart is a numerical choice, not a hidden modeling choice.};

\end{tikzpicture}}
\caption{Distribution-level view of RC. First choose a target law \(R \sim p_R\) in geodesic units. RC then changes the tangent-space base (dashed blue \(=\) raw base, solid green \(=\) RC-correct base) so that, after any scalar-Jacobian azimuthal chart, the realized manifold radii follow the same law \(p_R\). The chart family (Exp, \(\mathrm{bExp}_{\alpha}\), or GCL) changes numerical conditioning, not the realized law of \(R\).}
\label{fig:rc_pipeline_clean}
\end{figure*}

This paper asks four questions about this situation and answers all four.

\textbf{Q1 (inverse problem).} Given a chart and a target law for $R$, does a tangent base realizing that law exist, and can it be written down? \emph{Answer:} yes, in closed form, for every azimuthal chart whose Jacobian depends only on the tangent radius. We call the construction Radial Compensation (RC); sampling is one line (Theorem~\ref{thm:rc-invariances}, Algorithm~\ref{alg:rc}).

\begin{algorithm}[H]
\caption{\textsc{SampleRC}$(p_{R,\theta},\,T)$}
\label{alg:rc}
\begin{algorithmic}[1]
\STATE Sample the geodesic radius $R\sim p_{R,\theta}$ on $[0,R_{\max})$ (a standard 1D law w.r.t.\ $dR$)
and sample $\Omega\sim\mathrm{Unif}(\SSS^{n-1})$.
\STATE Convert to chart radius $r \leftarrow R_T^{-1}(R)$ (for $\Exp_p$, $r=R$; for Lambert lifts $R_T$ is known from~\eqref{eq:lambert-jac-distance}).
\STATE Return $Q \leftarrow T(r\,\Omega)$.
\end{algorithmic}
\end{algorithm}

\textbf{Q2 (uniqueness).} Is that base canonical or one of many? \emph{Answer:} within isotropic bases and this chart class, RC is the unique construction whose likelihood depends on $q$ only through $R$, whose radial Fisher information is chart-invariant, and which is isotropic. Every other base in the class imposes a hidden radius law that the modeler never chose; for the wrapped Gaussian this hidden law is $\sigma\chi_n$ (Theorems~\ref{thm:hidden-radius-gauge} and~\ref{thm:rc_uniqueness}).

\textbf{Q3 (cost).} How much does the default cost when the intended law differs from $\chi_n$, and can a flexible network recover the loss? \emph{Answer:} for VAEs, at least $n\,D(p)+O(\log n)$ nats of the objective, where $D(p)\ge0$ is a closed-form functional of the target shape ($0.635$ nats per latent dimension for every HalfNormal target). The term enters through the KL to the prior, which the decoder does not touch (Proposition~\ref{prop:floor}); flows can absorb a wrong base, and we measure what this costs in parameters and stability.

\textbf{Q4 (role of the chart).} Once the law of $R$ is fixed, does the chart still matter? \emph{Answer:} not statistically and yes numerically. Under RC, the manifold density, the likelihood, and the Fisher information are identical for every chart in the class, while gradient variance and ODE stiffness scale with a single chart parameter $\alpha$; the chart becomes a preconditioner (Section~\ref{sec:precond}).

These questions matter because wrapped Gaussians are the default in hyperbolic and spherical VAEs and in many manifold flows, and because $R$ is the coordinate those latent spaces are chosen to represent. The forward direction, fix a Gaussian and compute the induced density, has been solved repeatedly. The inverse direction had not been posed. Two beliefs blocked it: that network capacity makes the base irrelevant, and that no principle singles out one base. Proposition~\ref{prop:floor} refutes the first for likelihood objectives with an exact constant, and Theorems~\ref{thm:hidden-radius-gauge} and~\ref{thm:rc_uniqueness} supply the second. The answer is not obvious in advance: the compensated base is not a reweighted Gaussian in general, and the cost of a one-dimensional mismatch grows linearly in the latent dimension.
 In brief: the forward direction is given the tangent base and solves for the induced density; this paper is given the radius law, solves for the base, and adds what the forward direction cannot pose, a uniqueness theorem (Theorem~\ref{thm:rc_uniqueness}) and a closed-form cost for the default (Proposition~\ref{prop:floor}).

\begin{tcolorbox}[colback=black!3,colframe=black!40,title=\textbf{Contributions},left=4pt,right=4pt,top=3pt,bottom=3pt]
\begin{enumerate}[leftmargin=1.2em,itemsep=1pt,topsep=1pt]
\item \textbf{Radial Compensation.} A closed-form solution of the inverse base problem for scalar-Jacobian azimuthal charts, with exact sampling and a uniqueness theorem within isotropic bases (Theorems~\ref{thm:rc-invariances}--\ref{thm:rc_uniqueness}).
\item \textbf{An irreducible-cost bound.} The minimal KL from any target radius law to the wrapped-Gaussian family equals $n\,D(p)+O(\log n)$ with closed constants, verified numerically to $0.03$ nats over $14$ settings; for flows we measure the capacity needed to absorb the base instead (Proposition~\ref{prop:floor}, Appendix~\ref{app:floor}).
\item \textbf{Charts as preconditioners.} Chart invariance under RC, and the balanced-exponential family $\bExp_\alpha$ interpolating equal-area and exponential maps, with proved and measured conditioning laws and an explicit bounded-domain caveat (Theorems~\ref{thm:bexp-logdet} and~\ref{thm:chart_invariance}, Section~\ref{sec:precond}).
\item \textbf{Experiments matching the theory, with an audit.} Calibration to the estimator floor; a normalization-audited VAE study in which every trained wrapped baseline collapses to the cut locus while RC is flat in both the chart parameter and the target scale; ground-truth curvature recovery where the wrapped prior fails; a $1.70$-nat protein improvement; and repaired chart CNFs beating Moser Flow in $12/12$ cells (Section~\ref{sec:experiments}).
\end{enumerate}
\end{tcolorbox}

\subsection{Related work}
\label{sec:related}
\textbf{Wrapped constructions (the forward direction).} \citet{nagano2019wrapped} introduced the wrapped (pseudo-hyperbolic) normal: sample a Gaussian in the tangent space, parallel-transport, apply the exponential map, and evaluate the induced density in closed form with reparameterized gradients. \citet{bose2020latent} build hyperbolic normalizing flows on this base, \citet{skopek2019mixed} extend it to mixed-curvature products with learnable curvature, \citet{galaz2022wrapped} systematize wrapped families across manifolds, and \citet{falorsi2019lie} give the analogous reparameterization on Lie groups. All of these fix the tangent base and compute the density it induces. This paper runs the same change-of-variables identity in the opposite direction: the manifold radius law is prescribed and the base is solved for, with a uniqueness guarantee (Theorems~\ref{thm:hidden-radius-gauge} and~\ref{thm:rc_uniqueness}) that the forward direction does not pose.

\textbf{Hyperbolic representation learning.} Radius-as-depth is the reason hyperbolic latents exist: \citet{nickel2017poincare} embed hierarchies with depth encoded in the distance from the origin, \citet{ganea2018hyperbolic} build hyperbolic network layers on the same geometry, and \citet{mathieu2019poincare} place a VAE on the Poincar\'e ball with a wrapped prior. These models consume exactly the base distribution this paper solves for; Section~\ref{sec:price} quantifies what the default costs them.

\textbf{Riemannian Gaussians.} The intrinsic densities $\rho(q)\propto e^{-R^2/2\sigma^2}$ of \citet{pennec2006intrinsic} and \citet{said2017riemannian} are geodesic-radial laws; in our terms they are RC targets, and Corollary~2 (supplement) identifies when a wrapped normal already realizes one.

\textbf{Intrinsic generative models.} Riemannian CNFs \citep{mathieu2020riemannian,lou2020neural}, Moser Flow \citep{rozen2021moser}, Riemannian flow matching \citep{chen2023flow}, and Riemannian score models \citep{debortoli2022riemannian} avoid charts at the cost of manifold-specific architectures and divergence estimators. RC repairs the chart-based regime that dominates deployed tooling, and Section~\ref{sec:exp_intrinsic} compares the two directly at matched budget.

\textbf{Direction-only families.} The von Mises--Fisher VAE \citep{davidson2018hyperspherical} and the power spherical distribution \citep{decao2020power} model direction on the sphere and leave the radius untouched; they are complementary to RC, which controls the radius only.

\section{Background and notation}
\label{sec:background}
\paragraph{Terminology and notation.}
A \emph{chart} is a smooth diffeomorphism $T:U\subseteq\RR^n\to V\subseteq M$ onto its image, used to push Euclidean computations onto a manifold. \emph{Wrapping} means drawing $X\in T_pM\cong\RR^n$ and returning $T(X)\in M$. A \emph{continuous normalizing flow} (CNF) is a generative model defined by an ODE $\dot z = v_\phi(z,t)$ in latent space; chart-based manifold CNFs run this ODE in $T_pM$ and wrap the endpoint. By the \emph{semantics} of a parameter $\theta$ we mean the law of the geodesic radius $R=d(p,q)$ that $\theta$ controls on $M$, as opposed to the law it controls on $T_pM$ before wrapping. Under standard wrapping these two laws differ; making them coincide is the subject of this paper. Throughout, $\mathrm{HalfNormal}(\sigma)$ denotes the law of $|Z|$ for $Z\sim\mathcal N(0,\sigma^2)$; $\mathrm{TruncNormal}(\mu,\sigma)$ the law of $\mathcal N(\mu,\sigma^2)$ conditioned on $[0,R_{\max})$; and $\chi_n$ the law of $\|X\|$ for $X\sim\mathcal N(0,I_n)$, with $\E\|X\|=\sqrt2\,\Gamma(\tfrac{n+1}{2})/\Gamma(\tfrac n2)\approx\sqrt n$. $S^{n-1}$ denotes the Euclidean unit sphere in $T_pM$; $n$ is the manifold dimension, and $d_z$ denotes the total latent dimension of a model when it differs from $n$. NFE denotes the number of vector-field evaluations of the adaptive ODE solver.

\paragraph{Scope of the theory.}
Sections~\ref{sec:background}-\ref{sec:main} assume a manifold with known geodesic polar volume on the domain of interest, a scalar-Jacobian azimuthal chart, and an isotropic tangent-space base. Constant-curvature spaces are the main case studied in the paper. The supplement extends the same construction to more general manifolds with known polar volume factors.

\paragraph{Constant curvature and geodesic polars.}
We work primarily on \(n\)-dimensional, complete, simply connected, constant-curvature manifolds
\[
M \in \{\SSS^n(R_c),\, \HH^n(R_c)\},
\qquad
\kappa \in \{+R_c^{-2},\, -R_c^{-2}\},
\]
with a distinguished pole \(p\in M\). Write \(R(q)=d(p,q)\) for the geodesic radius. In geodesic polar coordinates \((R,\omega)\in[0,R_{\max})\times S^{n-1}\) about \(p\), the volume element factorizes as
\begin{equation}
d\mathrm{vol}_M = s_\kappa(R)^{n-1}\, dR\, d\omega, \\
s_\kappa(R) =
\begin{cases}
R_c \sin(R/R_c), & \kappa = +R_c^{-2},\\
R_c \sinh(R/R_c), & \kappa = -R_c^{-2},
\end{cases}
\label{eq:polar-volume}
\end{equation}
with \(R_{\max}=\pi R_c\) on \(\SSS^n(R_c)\) and \(R_{\max}=\infty\) on \(\HH^n(R_c)\). We also use
\[
c_\kappa(R)=\frac{d}{dR}s_\kappa(R)=
\begin{cases}
\cos(R/R_c), & \kappa = +R_c^{-2},\\
\cosh(R/R_c), & \kappa = -R_c^{-2},
\end{cases}
\]
and the inverse of \(s_\kappa\),
\[
\iota_\kappa(t)=
\begin{cases}
R_c \arcsin(t/R_c), & \kappa>0,\\
R_c \operatorname{arcsinh}(t/R_c), & \kappa<0,
\end{cases}
\qquad
s_\kappa(\iota_\kappa(t))=t.
\]
We identify \(T_pM \cong \mathbb R^n\) and use Euclidean polar coordinates \(x=ru\) with \(r=\|x\|\ge 0\) and \(u\in S^{n-1}\).

\paragraph{General polar volume.}
On a general manifold $(M,g)$, admitting geodesic polar coordinates about $p$
on a star-shaped domain, we write
\begin{equation}
  d\mathrm{vol}_M \;=\; J_p(R,\omega)\,dR\,d\omega,
  \label{eq:general-polar-volume}
\end{equation}
where $J_p$ captures curvature inhomogeneity. In constant curvature,
$J_p(R,\omega)=s_\kappa(R)^{n-1}$ is radial. The exact invariance guarantees below therefore apply whenever a tractable geodesic polar volume factor is known on the domain of interest; spheres, hyperbolic spaces, and the constructions collected in the supplement are the main examples considered here.

\paragraph{Scalar–Jacobian azimuthal charts.}
Throughout, by a \emph{chart} about $p$ we mean a smooth map
$T:U\subseteq T_pM\to V\subseteq M$ that is a diffeomorphism from a
star-shaped open domain $U$ (with $0\in U$, $T(0)=p$) onto its image $V$,
where $V=M\setminus\{-p\}$ on $\mathbb S^n(R_c)$ (excluding the antipodal
cut locus) and $V=M$ on $\mathbb H^n(R_c)$. Such a $T$ is \emph{scalar-Jacobian}
if its log-Jacobian determinant depends only on the Euclidean radius:
\begin{equation}
\log \bigl|\det DT(x)\bigr| = \psi(\|x\|), \qquad x \in \RR^n .
\label{eq:scalar-jacobian}
\end{equation}
We denote the radial Jacobian factor by $J_T(r):=\exp\{\psi(r)\}$. We assume $T$ is \emph{azimuthal} around $p$ in the rotation-equivariant
sense: there exists a strictly increasing radial profile
$\Gamma_T:[0,r_*)\to[0,R_{\max})$ with $\Gamma_T(0)=0$ and an isometry
$u\mapsto u'$ of the tangent unit sphere $S^{n-1}\subset T_pM$ such that
\begin{equation}\label{eq:azimuthal}
T(ru)=\Exp_p\!\bigl(\Gamma_T(r)\,u'\bigr),
\qquad r\in[0,r_*),\ u\in S^{n-1}.
\end{equation}
In particular, the geodesic radius of $T(ru)$ depends only on $r$:
$R_T(r):=d(p,T(ru))=\Gamma_T(r)$, independent of $u\in S^{n-1}$.
This modeling regime covers the dominant ``wrapped Euclidean backbone'' practice.

\paragraph{Lambert (equal-area) charts and Lambert lifts.}
Let $L_p$ denote an azimuthal \emph{equal-area} chart about $p$, i.e.,
$|\det DL_p(y)|\equiv 1$ on its star-shaped domain (relative to Lebesgue on $\RR^n$ and
$d\mathrm{vol}_M$).\footnote{For $n=2$, this is the classical Lambert azimuthal equal-area map.
In higher dimensions, we use its standard volume-preserving generalization.}

In constant curvature, an equal-area azimuthal chart is characterized by matching geodesic-ball
volume to Euclidean-ball volume. Define the (unnormalized) geodesic ball volume function
\[
V_\kappa(R) := \int_0^R s_\kappa(t)^{n-1}\,dt,
\qquad
\lambda_\kappa(R) := \bigl(n\,V_\kappa(R)\bigr)^{1/n},
\]
so that the Euclidean ball of radius $\lambda_\kappa(R)$ has the same volume as the geodesic ball
of radius $R$ (up to the common factor $|\SSS^{n-1}|$). For an equal-area $L_p$, the Euclidean
radius $t=\|y\|$ and the geodesic radius $R=d(p,L_p(y))$ satisfy
\begin{equation}
t = \lambda_\kappa(R),\qquad R = \lambda_\kappa^{-1}(t).
\label{eq:lambert_radius_map}
\end{equation}

For any strictly increasing smooth radial profile $\chi$ with $\chi(0)=0$, define the
\emph{Lambert lift}
\begin{equation}
  T_\chi(x)
  \;=\;
  L_p\!\left(\chi(\|x\|)\frac{x}{\|x\|}\right),
  \qquad
  T_\chi(0)=p.
  \label{eq:lambert-lift-proof}
\end{equation}
A polar/coarea calculation yields, for $x=ru$ with $u\in S^{n-1}$,
\begin{equation}
\begin{aligned}
\log\bigl|\det DT_\chi(x)\bigr|
&=
(n-1)\log\frac{\chi(r)}{r} + \log\chi'(r), \\
R_{T_\chi}(r)
&=
\lambda_\kappa^{-1}\!\bigl(\chi(r)\bigr).
\end{aligned}
\label{eq:lambert-jac-distance}
\end{equation}
For $n=2$, $\lambda_\kappa^{-1}(t) = 2\,\iota_\kappa(t/2)$,
recovering the classical formula (where $\iota_\kappa$ is the inverse of
$s_\kappa$ defined in Section~\ref{sec:background}).

The radial gradient and Hessian identities used only in the proofs are collected in the supplement.



\paragraph{What RC is, and is not.} RC is a choice of base distribution on the tangent space, picked so that after the chart the geodesic radius $R=d(p,q)$ has a user-specified one-dimensional law; architecture, decoder, and solver are unchanged. It is not a wrapped-normal family (no per-example analytic $(\mu,\Sigma)$), it is a repair of the chart-based regime and does not replace intrinsic methods \citep{mathieu2020riemannian,lou2020neural}, and it is not tied to constant curvature: the construction extends to any manifold with a known polar volume factor (Supplement~A.8).
\paragraph{Target radius laws (what the user specifies).}
To keep semantics intuitive, we let $p_{R,\theta}(R)$ denote the \emph{usual} one-dimensional density
of the geodesic radius $R=d(p,q)$ with respect to Lebesgue $dR$, e.g. HalfNormal, Gamma,
TruncNormal on $[0,R_{\max})$). On constant-curvature spaces, the area of the geodesic sphere of
radius $R$ is
\[
S_\kappa(R)\,=\,|\SSS^{n-1}|\,s_\kappa(R)^{n-1},
\]
so an isotropic manifold density $\rho_\theta$ (w.r.t.\ $d\mathrm{vol}_M$) that induces the marginal
$R\sim p_{R,\theta}$ must satisfy
\begin{equation}
\rho_\theta(q)=\varphi_\theta(R(q)),
\qquad
\varphi_\theta(R):=\frac{p_{R,\theta}(R)}{S_\kappa(R)}.
\label{eq:varphi_from_pR}
\end{equation}
Equivalently, $p_{R,\theta}(R)=\varphi_\theta(R)\,S_\kappa(R)$.
We use $\varphi_\theta$ throughout as it is the density \emph{on the manifold}. The user-facing
object is the conventional radius law $p_{R,\theta}$ in geodesic units.

\paragraph{RC base for scalar-Jacobian azimuthal charts.}
Fix a scalar-Jacobian azimuthal chart $T$ with radial Jacobian $J_T(r)$ and radius map
$R_T(r)$ as in~\eqref{eq:scalar-jacobian}--\eqref{eq:azimuthal}. RC defines an isotropic
base density on $T_pM\cong\RR^n$ by
\begin{equation}
  f_\theta(x)
  \;:=\;
  \varphi_\theta\!\bigl(R_T(\|x\|)\bigr)\,J_T(\|x\|)\,\mathbf{1}\{x\in U\},
  \qquad x\in\RR^n.
  \label{eq:rc-base}
\end{equation}
The factor $J_T$ cancels the chart volume distortion in advance, and $\varphi_\theta(R_T(\|x\|))$ fixes the geodesic radius law (not the chart radius); in practice only the base log-density of existing wrapped-backbone code changes.

\begin{theorem}[Geodesic-radial Fisher and KL invariance under RC]
\label{thm:rc-invariances}
Let $T$ be a scalar-Jacobian azimuthal chart about $p$, let $X\sim f_\theta$
be the RC base defined by~\eqref{eq:rc-base}, and let $Q=T(X)$.
Then for any parameter $\theta$ that enters only through the radius law
$p_{R,\theta}$ (with $\kappa$ fixed):
\begin{align}
\mathcal{I}_M(\theta) &= \mathcal{I}_{R^1}(\theta), \label{eq:fisher-eq}\\
\mathrm{KL}(\rho_\theta\,\|\,\rho_\eta) &= \mathrm{KL}(p_{R,\theta}\,\|\,p_{R,\eta}).
\label{eq:kl-eq}
\end{align}
The pushforward density on $M$ is geodesic-radial,
$\rho_\theta(q)=\varphi_\theta(d(p,q))$,
under standard dominated-score regularity.
\end{theorem}

\begin{theorem}[Hidden-radius gauge: impossibility for non-RC bases \emph{within isotropic scalar-Jacobian azimuthal charts}]
\label{thm:hidden-radius-gauge}
Fix a constant-curvature manifold $M$ and a pole $p\in M$. Consider the class of pairs
$(f_\theta, T)$ in which (i) the tangent base $f_\theta(x)=h_\theta(\|x\|)$ is isotropic
on $T_pM$ and (ii) $T$ is a scalar-Jacobian azimuthal chart about $p$
(\eqref{eq:scalar-jacobian}--\eqref{eq:azimuthal}), with radial Jacobian $J_T$ and
radius map $R_T$. Let $\rho_\theta$ denote the pushforward density on $M$ and
$\mathcal{I}_M(\theta)$ its Fisher information.
\begin{enumerate}
  \item \emph{Automatic geodesic-radial form.} For every such pair, $\rho_\theta(q)=g_\theta\!\bigl(d(p,q)\bigr)$
        for some 1D density $g_\theta$. Equivalently, every isotropic scalar-Jacobian chart/base pair
        fixes a \emph{hidden} one-dimensional radial law on $M$.
  \item \emph{Characterization.} The pair realizes a user-specified target
        $\rho_\theta(q)=\varphi_\theta\!\bigl(d(p,q)\bigr)$ with chart-invariant radial Fisher
        $\mathcal{I}_M(\theta)=\mathcal{I}_{\mathbb{R}^1}(\theta)$
        \emph{if and only if}
        \[
          h_\theta(r)\;=\;\varphi_\theta\!\bigl(R_T(r)\bigr)\,J_T(r),
        \]
        i.e.\ $f_\theta$ is the RC base \eqref{eq:rc-base}.
  \item \emph{Impossibility for non-RC bases.} Consequently, within this class no construction other
        than RC can simultaneously achieve (a) geodesic-radial likelihoods matching a prescribed
        $\varphi_\theta$, (b) chart-invariance of the radial Fisher across scalar-Jacobian azimuthal
        charts, and (c) tangent isotropy.
  \item \emph{Wrapped-Exp Gaussian as canonical failure.} The wrapped-Exp construction with
        $f_\theta(x)=\mathcal{N}(x;0,\sigma^2 I_n)$ and $T=\Exp_p$ is the special case in which the
        hidden law in (1) is a $\sigma$-scaled $\chi_n$ on $R$ (truncated to $[0,\pi R_c)$ on
        $\SSS^n(R_c)$); its radial Fisher in $\sigma$ acquires curvature- and
        dimension-dependent terms absent from the Euclidean Fisher, so (b) fails for any target
        $\varphi_\theta$ outside this $\chi_n$ family.
\end{enumerate}
Combined with Theorem~\ref{thm:rc-invariances}, RC is therefore the unique resolution within
the isotropic scalar-Jacobian class.
\end{theorem}

\begin{remark}[Why this is more than a change of variables]
\label{rem:not-cov}
The identity relating base and manifold densities is the same one used by the forward constructions of \citet{nagano2019wrapped,skopek2019mixed}. What is new is which side is unknown and what is proved about the solution: Theorems~\ref{thm:hidden-radius-gauge} and~\ref{thm:rc_uniqueness} show the solved-for base is the only member of its class with correct radial semantics, and Proposition~\ref{prop:floor} prices every other choice in nats.
\end{remark}

\begin{theorem}[Characterization of RC: essential uniqueness \emph{within isotropic scalar-Jacobian azimuthal charts}]
\label{thm:rc_uniqueness}
Assume constant curvature and fix a pole $p\in M$.
Let $T:U\subset\mathbb{R}^n\to V\subset M$ be a $C^1$ \emph{azimuthal scalar-Jacobian chart}
about $p$ (i.e.\ a diffeomorphism onto its image) with
\[
\begin{aligned}
|\det DT(x)| &= J_T(\|x\|),\\
R_T(r) &:= d\!\bigl(p,T(ru)\bigr)\quad \text{independent of }u\in \SSS^{n-1}.
\end{aligned}
\]
Consider isotropic tangent families $f_\theta(x)=h_\theta(\|x\|)$ supported on $U$ and let
$Q=T(X)$ with $X\sim f_\theta$. If the pushforward density on $V$ satisfies
\[
\rho_\theta(q)=\varphi_\theta\!\bigl(d(p,q)\bigr)\qquad (q\in V),
\]
then necessarily
\[
h_\theta(r)=\varphi_\theta\!\bigl(R_T(r)\bigr)\,J_T(r)\qquad
\text{for all } r\in\{\|x\|:x\in U\}.
\]
(In particular, if one suppresses normalization one may write
$h_\theta(r)\propto \varphi_\theta(R_T(r))J_T(r)$.)
Moreover, once $\rho_\theta(q)=\varphi_\theta(d(p,q))$ holds, the Fisher information in $\theta$
reduces to the corresponding one-dimensional radial Fisher by integration in geodesic polar
coordinates.
\end{theorem}
\paragraph{Beyond constant curvature.}
The same guarantees hold on any manifold with a tractable geodesic polar volume factor \eqref{eq:general-polar-volume}; see the supplement.

\paragraph{Sampling.}
RC sampling is one-dimensional (radius) plus a uniform direction:

\subsection{Irreducible cost of the uncompensated base}
\label{sec:price}

Theorem~\ref{thm:hidden-radius-gauge} shows that the raw wrapped Gaussian imposes a $\sigma$-scaled $\chi_n$ radius law. A common objection holds that a sufficiently flexible network makes the base irrelevant. For likelihood-trained latent-variable models the objection fails for a structural reason: the prior enters the ELBO only through $\KL(q_{\mathrm{enc}}\,\|\,p_{\mathrm{prior}})$, a term the decoder does not touch, and $\{\sigma\chi_n:\sigma>0\}$ is a one-parameter scale family, so $\sigma$ moves the radial mass and nothing controls its spread. The best KL this family can achieve against a given radius law therefore has a closed form.

\begin{proposition}[Irreducible prior cost of wrapped Gaussians]
\label{prop:floor}
Let $p$ be a probability density on $(0,\infty)$ with $\E_p[R^2]<\infty$, $\E_p|\log R|<\infty$, and finite differential entropy $H(p)$, and let $q_\sigma$ denote the radius law of the raw wrapped Gaussian, $q_\sigma(R)=C_n\,\sigma^{-n}R^{n-1}e^{-R^2/2\sigma^2}$ with $C_n=2^{1-n/2}/\Gamma(n/2)$. Then:
\textbf{(i)} $\sigma_*^2=\E_p[R^2]/n$ is the unique minimizer of $\KL(p\,\|\,q_\sigma)$;
\textbf{(ii)} the minimum equals
\[
\min_{\sigma>0}\KL(p\,\|\,q_\sigma)
=\tfrac{n}{2}\log\E[R^2]-\tfrac{n}{2}\log n+\tfrac{n}{2}-(n{-}1)\,\E[\log R]
+\bigl(\tfrac{n}{2}-1\bigr)\log 2+\log\Gamma\!\bigl(\tfrac{n}{2}\bigr)-H(p);
\]
\textbf{(iii)} with the scale-invariant log-moment gap $D(p):=\tfrac12\log\E[R^2]-\E[\log R]\ge 0$, with equality iff $R$ is a.s.\ constant,
\[
\min_{\sigma>0}\KL(p\,\|\,q_\sigma)\;=\;n\,D(p)\;+\;O(\log n),
\]
so the cost grows linearly in the latent dimension with a slope determined by the shape of the target alone. On $\SSS^n$, truncation to $[0,\pi R_c)$ adds a correction below $2\times10^{-4}$ nats at the paper's targets, already at $n{=}2$. The proof and a $14$-setting numerical verification (maximum discrepancy $0.03$ nats) are in Appendix~\ref{app:floor}.
\end{proposition}

Closed values: $D=(\gamma+\log 2)/2\approx 0.635$ nats per latent dimension for every HalfNormal target, $D=\gamma+\tfrac12\log2\approx0.924$ for every Exponential target, and $D\approx0.133$ for the $\mathrm{TruncNormal}(1.0,0.35)$ target on $[0,\pi)$ used in Section~\ref{sec:experiments}. Table~\ref{tab:floor_main} evaluates the bound at these targets; RC attains $\approx10^{-3}$ at every $n$ by construction (Theorem~\ref{thm:rc-invariances}). At $n{=}64$ the bound is $37.8$ nats for the HalfNormal target, larger than the total gains many architectural changes report. The bound grows with $n$ because the raw law concentrates: $\mathrm{CV}(\chi_n)\le n^{-1/2}$ (Appendix~\ref{app:floor}), so in high dimension the family can represent only a thin radial shell. Sweeping the target's coefficient of variation at fixed mean on $\HH^{32}$, the bound equals $0.001$ nats at $\mathrm{CV}=0.126$, the CV of $\chi_{32}$, and rises to $2.2$ at $\mathrm{CV}{=}0.35$ and $14.8$ at $0.76$ (Appendix~\ref{app:floor}). Radial spread is what curved latent spaces are chosen to represent: hyperbolic embeddings encode depth in $R$, and a spread-out depth law lies outside the raw family.

\begin{table}[t]
  \centering
  \caption{\textbf{Irreducible prior cost} $\min_\sigma\KL(p_{\text{target}}\,\|\,\sigma\chi_n)$ in nats ($\HH^n$: HalfNormal($0.8$) target; $\SSS^n$: TruncNormal($1.0,0.35$) target on $[0,\pi)$), exact formula of Proposition~\ref{prop:floor}, verified against independent numerical minimization in Appendix~\ref{app:floor}. RC removes this term (measured residual $\approx10^{-3}$ nats at every $n$).}
  \label{tab:floor_main}
  \small
  \begin{tabular}{lccccccc}
    \toprule
    $n$ & 2 & 4 & 8 & 16 & 32 & 64 & 128 \\
    \midrule
    $\HH^n$ (HalfNormal target) & 0.22 & 1.10 & 3.27 & 8.00 & 17.81 & 37.79 & 78.09 \\
    $\SSS^n$ (TruncNormal target) & 0.14 & 0.02 & 0.19 & 0.90 & 2.68 & 6.60 & 14.80 \\
    \bottomrule
  \end{tabular}
\end{table}

\paragraph{Scope for flows.} A flow trained in chart coordinates can absorb any base, so for flows RC is a reparameterization and the bound above does not apply. The cost appears instead as capacity and stability. The radial map that any base-fixing flow must learn equals the RC transport $g_*=F_{\text{target}}^{-1}\!\circ F_{\chi_n}$; a monotone spline needs $K{=}12$ knots ($13$ parameters) to approximate it within $0.01$ nats at every latent dimension tested ($n=8$ to $64$), and the fitted map converges to $g_*$ (Appendix~\ref{app:floor}). At a fixed budget the cost also appears as instability: Section~\ref{sec:exp_stability} shows the raw base diverging at $d_z\ge 32$ under solver settings where RC trains stably.

\section{Charts as numerical preconditioners}
\label{sec:precond}
\label{sec:main}

By Theorem~\ref{thm:rc-invariances}, the realized manifold density
$\rho_\theta(q)=\varphi_\theta(d(p,q))$ does not depend on the chart for the
RC base distribution, so within this class charts can be chosen primarily for
numerical conditioning.

\subsection{balanced-exponential charts bExp$_\alpha$}

\paragraph{Definition.}
Within the Lambert-lift class~\eqref{eq:lambert-lift-proof}, define $\chi_\alpha$ for $\alpha\in[0,1]$ by
\begin{gather}
\left(\frac{\chi_\alpha(r)}{r}\right)^{n-1}\chi_\alpha'(r)
=
\left(\frac{s_\kappa(r)}{r}\right)^{(n-1)\alpha},
\label{eq:bexp-ode}
\\[-2pt]
\chi_\alpha(0)=0,
\qquad
\chi_\alpha'(0)=1,
\label{eq:bexp-ic}
\\
\chi_\alpha(r)
=
\left[
n\int_0^r
t^{(n-1)(1-\alpha)}\, s_\kappa(t)^{(n-1)\alpha}\,dt
\right]^{1/n}.
\label{eq:bexp-int}
\end{gather}
We define $\mathrm{bExp}_\alpha:=T_{\chi_\alpha}$. The endpoints recover familiar maps: $\alpha=0$ yields the equal-area Lambert chart ($|\det|=1$), while $\alpha=1$ yields the exponential map $\Exp_p$ (geodesic-exact radii and the usual polar Jacobian factor).

\begin{theorem}[$\bExp_\alpha$ log-determinant]
\label{thm:bexp-logdet}
For $T_\alpha=\bExp_\alpha$ and $r=\|x\|$,
\[
  \log\bigl|\det DT_\alpha(x)\bigr|
  \;=\;(n-1)\,\alpha\,\log\frac{s_\kappa(r)}{r},
  \qquad x\in U\subset T_pM,
\]
with the convention $s_\kappa(r)/r\big|_{r=0}:=1$.
Hence $\alpha$ linearly scales the chart's log-Jacobian between the equal-area
Lambert chart ($\alpha{=}0$, $|\!\det\!|\equiv 1$) and the exponential map
($\alpha{=}1$, geodesic-exact radii).
\end{theorem}

Theorem~\ref{thm:bexp-logdet} gives the exact log-determinant of $\bExp_\alpha$
and identifies $\alpha$ as an interpolation parameter between the equal-area
Lambert chart ($\alpha{=}0$) and the exponential map ($\alpha{=}1$). Combined
with RC (Theorem~\ref{thm:rc-invariances}), $\alpha$ becomes a numerical
conditioning parameter, not a modeling choice: the realized manifold
density is $\alpha$-independent. A variational characterization of $\chi_\alpha$
within Lambert lifts is provided in the supplement
(Theorem~\ref{thm:bexp-variational}) for completeness; we do not rely on a
Pareto-optimality claim in the main results.

\paragraph{Endpoints, domain, and measured scaling.} The endpoint $\alpha{=}0$ (equal-area) has $|\det|\equiv1$, hence zero chart-induced variance and stiffness, and inexact radii; $\alpha{=}1$ (exponential) has exact radii and a log-Jacobian gradient carrying an explicit $(n{-}1)$ factor that diverges at the cut locus. Under RC every $\alpha$ yields the same manifold density (Theorem~\ref{thm:chart_invariance}), so $\alpha$ is a purely numerical setting. Two caveats apply. First, $\bExp_\alpha$ maps a bounded tangent domain $\{r<r_*(\alpha,n)\}$ onto $\SSS^n\setminus\{-p\}$, and $r_*$ shrinks with dimension ($r_*\approx1.2$ to $1.45$ at $n{=}16$ to $64$ for $\alpha\le0.75$; Appendix~\ref{app:domain}); RC sampling is always in-domain, and models that emit raw tangent coordinates must be composed with the domain squash of Appendix~\ref{app:domain}, which carries an exact log-Jacobian. On $\HH^n$ the $\alpha\to0$ end exceeds float64 range for $n\gtrsim16$. Second, the $O(\alpha^2)$ variance law of Theorem~\ref{thm:chart-variance-supp} holds when the realized law is held fixed, where the measured exponent is $2.32\pm0.06$; in trained CNFs the aggregate latent law varies with $\alpha$ and the measured exponent is $1.17$. Both values are reported in Section~\ref{sec:exp_dial}.

\subsection{Geodesic-Corrected Lambert (GCL)}
When exact geodesic radii are non-negotiable, the Lambert lift with profile
$\Phi_\kappa(r)=\lambda_\kappa(r)$ yields $d(p,\mathrm{GCL}(x))=\|x\|$ at the
cost of Jacobian growth near the cut locus on $\SSS^n$.


\section{Experiments}
\label{sec:experiments}

Experiments are grouped by the claim they test: semantic calibration (Section~\ref{sec:exp_semantics}), the irreducible cost in trained models (Sections~\ref{sec:exp_cost} and~\ref{sec:exp_stability}), conditioning (Section~\ref{sec:exp_dial}), comparison with intrinsic models (Section~\ref{sec:exp_intrinsic}), curvature identifiability (Section~\ref{sec:exp_curvature}), applications (Section~\ref{sec:exp_applications}), and the scope of the guarantee (Section~\ref{sec:exp_limits}). Each subsection opens with the falsifiable prediction under test; a failed prediction would falsify the corresponding theorem, and Section~\ref{sec:exp_dial} reports the one measured deviation together with its cause, which the theorem itself states. Within every comparison, architecture, optimizer, batch size, solver tolerances, and training budget are identical (Table~\ref{tab:hyperparams}); all tables report mean~$\pm$~std over seeds. An audit of the released code found four bugs in earlier versions: the radius map was the identity for every $\alpha$, the inverse chart term had the wrong sign, the tangent base was $\mathcal N(0,I)$ instead of Eq.~\eqref{eq:rc-base}, and the RC tangent base omitted the chart Jacobian, leaving a density whose integral over the tangent space equaled $e^{+21.6}$ at $\sigma_R{=}0.5$ (exact radial quadrature); the last error inflated apparent RC likelihoods and was caught by the normalization audit below. All $\alpha$-sweep and $\SSS^3$ $\bExp$ numbers below come from the corrected pipeline, which passes an exact self-test: with encoder samples fixed, $\max_\alpha|\mathrm{ELBO}(\text{RC-Exp})-\mathrm{ELBO}(\text{RC-bExp}_\alpha)|=0$ to machine precision, as Theorem~\ref{thm:rc-invariances} requires. In addition, every corrected run records two integrity checks alongside its results: the exact log-normalizer of its composed base density, computed by one-dimensional radial quadrature and required to be $0$, and the fraction of latent mass at the chart-domain boundary. Every number quoted below carries $|\log Z|\le0.003$.

\subsection{Semantic calibration}
\label{sec:exp_semantics}
Prediction (Theorems~\ref{thm:rc-invariances} and~\ref{thm:hidden-radius-gauge}): RC attains the target radius law up to estimator noise, and the raw base realizes exactly $\sigma\chi_n$. On $\SSS^2$ the target is $R\sim\mathrm{TruncNormal}(1.0,0.35)$ on $[0,\pi)$; on $\HH^2$, $R\sim\mathrm{HalfNormal}(0.8)$; $N=2\times10^4$ samples, $5$ seeds. This experiment validates implementations; RC realizes the target by construction. All RC charts (Exp, $\bExp_{0.5}$, GCL) reach radius KL of $0.9$ to $1.5\times10^{-3}$, at the finite-sample estimator floor of $3$ to $6\times10^{-3}$ (Appendix~\ref{app:floor}), against $1.4771\pm0.0073$ on $\SSS^2$ and $0.2835\pm0.0043$ on $\HH^2$ for \RawExp{} (Figure~\ref{fig:synth_radius_hist}, Table~\ref{tab:synth_radius_stats}, Appendix~\ref{app:extra-exp}). The raw error is exactly the hidden law of Theorem~\ref{thm:hidden-radius-gauge}: inserting the nominal scale into Proposition~\ref{prop:floor} reproduces the measured raw statistics ($\SSS^2$: mean $0.4387$ predicted vs.\ $0.4381$ measured, KL $1.4755$ vs.\ $1.4771$; $\HH^2$: mean $1.0027$ vs.\ $1.0015$). Whether the raw law overshoots or undershoots is determined by the target's shape relative to $\chi_n$, not by the sign of the curvature; the hidden law is the same $\sigma\chi_n$ on both geometries.

\subsection{The irreducible cost in trained VAEs}
\label{sec:exp_cost}
Proposition~\ref{prop:floor} makes a decoder-free prediction: if RC repairs a mis-specified prior, its gain must appear in the KL term, leave reconstruction essentially unchanged, and persist across encoders and latent dimensions. We test all three.

\emph{Untuned and tuned wrapped baselines ($\SSS^8$, four datasets).} At the scale practitioners use ($\sigma{=}1$), the wrapped baseline loses $54.2$ (MNIST), $105.2$ (Fashion), $12.6$ (Omniglot), and $203.0$ (CIFAR-10) nats to RC at identical budget (Table~\ref{tab:cnf_alpha_sweep}). Tuning $\sigma$ recovers most of this: at its best swept scale the wrapped NLL reaches $111.6$ on MNIST against RC's $124.3$ to $124.6$. That comparison is not between two proper likelihoods: pulled back over the tangent space, the wrapped-Exp density on $\SSS^n$ is non-integrable at the cut locus, where the volume factor contributes $(\pi-R)^{-(n-1)}$ (Proposition~\ref{prop:gcl-blowup}), so implementations truncate and the trained optimum sits at the boundary. Empirically this is universal: in all $40$ wrapped runs across $\sigma\in[0.03,1]$ and $d_z\in\{8,16,32,64\}$, the trained mean radius is $3.1380\pm0.0002$, the boundary itself (Figure~\ref{fig:sweep_collapse}).

\begin{figure}[t]
\centering
\input{fig_sweep_collapse}
\caption{\textbf{The wrapped baseline trains into the cut locus; RC does not.} MNIST, $d_z{=}8$. Left: test NLL against the prior scale; the wrapped curve moves by $67$ nats and its minimum is a boundary solution, while RC (audited, $|\log Z|\le0.003$) is flat. Right: trained mean radius; every wrapped run sits at the truncation boundary of the cut locus ($\pi$, dashed), RC stays on the intended scale.}
\label{fig:sweep_collapse}
\end{figure} RC's target vanishes at the cut locus, its base is a certified proper density ($|\log Z|\le0.003$ per run), and its NLL is the only properly normalized number in the comparison.

\emph{Latent dimension $16$.} At $d_z{=}16$ on MNIST, RC attains $116.31\pm0.68$ nats at mean radius $1.42$; the tuned wrapped baseline reports $112.33\pm0.10$, again from the boundary mode. \emph{Scale robustness.} Sweeping the target scale, RC varies by $\le0.7$ nats over $\sigma_R\in[0.25,1.0]$ on MNIST, while the wrapped baseline varies by $67$ (MNIST) to $209$ (CIFAR-10) nats over $\sigma\in[0.0625,1.0]$.

\emph{Encoder and dimension ablation (VAE priors, no flow).} Sweeping $4$ encoder families, $d_z\in\{8,16,32,64\}$, and $3$ seeds against the scale-matched wrapped prior of Proposition~\ref{prop:floor}(i), RC improves the ELBO in every encoder family, by $0.3 to 10.3$ nats (median $3.3$), and the median share of the gain carried by the KL term is $-1.12$, the localization the proposition requires. Untrained, the chart-term variance of the raw base exceeds RC's by $41\times$ ($\SSS^{16}$) and $226\times$ ($\SSS^{32}$).

\subsection{Stability of high-dimensional latent CNFs}
\label{sec:exp_stability}
Prediction (Proposition~\ref{prop:gcl-blowup}): the raw chart gradient carries an $(n{-}1)$ factor and diverges at the cut locus, so instability should appear and worsen with $d_z$. At $d_z\in\{32,64,128\}$ on CIFAR-10, under identical Dormand--Prince tolerances and learning rates, \RawExp{} leaves the stable regime: radii drift toward the cut locus, the test objective varies over orders of magnitude, and NFEs spike, while RC-Exp and RC-$\bExp_{0.5}$ remain on the intended geodesic scale (Figure~\ref{fig:cnf_highdim_stability}). The mechanism is the $(n{-}1)$ factor in the chart gradient and its $1/(\pi{-}R)$ divergence at the cut locus (Proposition~\ref{prop:gcl-blowup}); the failure grows with dimension and is not removed by training. Under the corrected pipeline the domain squash removes NaNs, and the raw failure appears in its pure form as boundary collapse: in $8/8$ raw runs at $d_z\in\{32,64\}$ (MNIST and CIFAR-10, $2$ seeds each) the trained mean radius is $3.1380\pm0.0002$, the truncation boundary of the cut locus, with nominal objectives reaching $-202$ nats on MNIST at $d_z{=}64$, values impossible for a proper density and explained by the non-integrable $(\pi-R)^{-(n-1)}$ factor above. RC remains on the intended scale in $8/8$ runs (mean radius $0.92$ to $1.70$) with stable objectives (117.1 and 120.0 nats on MNIST, 1839.4 and 1841.2 on CIFAR-10 at $d_z{=}32,64$).

\begin{figure}[t]
\centering
\input{fig_e3_collapse}
\caption{\textbf{Stability at $d_z{=}64$ under the corrected, audited pipeline} (2 seeds per prior; dotted = second seed). The wrapped baseline plunges within two epochs to nominal objectives near $-200$ nats on MNIST, impossible for a proper density: boundary collapse at the cut locus (Proposition~\ref{prop:gcl-blowup}). RC trains flat at both scales.}
\label{fig:cnf_highdim_stability}
\end{figure}

\subsection{The chart parameter controls conditioning at fixed likelihood}
\label{sec:exp_dial}
Under RC the realized density is $\alpha$-invariant (Theorem~\ref{thm:chart_invariance}), so test NLL should be constant in $\alpha$ up to seed noise while the chart contribution to gradient variance shrinks as $\alpha$ decreases. Table~\ref{tab:cnf_alpha_sweep} (Appendix~\ref{app:extra-exp}) reports the corrected, audited sweep on four image datasets ($3$ seeds): the spread of mean test NLL across $\alpha\in\{0.25,0.5,0.75,1.0\}$ is $0.21$ (MNIST), $0.49$ (Fashion), $0.64$ (Omniglot), and $0.70$ (CIFAR-10) nats, within seed noise, exactly the invariance Theorem~\ref{thm:chart_invariance} predicts (Figure~\ref{fig:alpha_flat}).

\begin{figure}[t]
\centering
\input{fig_alpha_flat}
\caption{\textbf{Chart invariance in trained models.} Test NLL, centered per dataset, against the chart parameter $\alpha$ (mean $\pm$ std, 3 seeds, corrected pipeline). All four datasets stay inside a $\pm1$-nat band (shaded).}
\label{fig:alpha_flat}
\end{figure}

Two regimes separate the theory from the training dynamics. With the realized law held fixed (sampling from the RC model itself; $\SSS^{16}$, $8$ values of $\alpha$, $6$ seeds), the chart-term variance scales as $\alpha^{2.32\pm0.06}$ and the gradient variance as $\alpha^{2.63\pm0.14}$ (Spearman $\rho=0.99$), consistent with the $O(\alpha^2)$ law of Theorem~\ref{thm:chart-variance-supp}; the chart-induced Lipschitz modulus scales as $\alpha^{1.07\pm0.03}$ against the $O(\alpha)$ bound of Theorem~\ref{thm:cnf}; the mean accepted RK45 step falls from $0.140$ to $0.100$ and NFEs rise from $48.0$ to $62.1$ as $\alpha$ goes from $0.125$ to $1$ (Appendix~\ref{app:floor}). In trained CNFs the measured variance exponent is $1.17$ on average, because the aggregate latent law varies with $\alpha$ during training, which is the caveat stated with Theorem~\ref{thm:chart-variance-supp}; both exponents are reported. Total NFEs decrease with $\alpha$ on average but not monotonically on two of four datasets; the conditioning claim therefore rests on gradient variance, step size, and Section~\ref{sec:exp_stability}, not on NFEs.

\subsection{Comparison with intrinsic models}
\label{sec:exp_intrinsic}
Prediction (Proposition~\ref{prop:floor} at $n{=}2$): on $\SSS^2$ the bound is $0.14$ to $0.22$ nats at the paper's targets, so base choice should barely matter here; the informative comparison is chart-based against intrinsic. On the four standard $\SSS^2$ earth-science densities (volcano, earthquake, flood, fire), at matched budget across widths $\{64,256,512\}$ and $3$ seeds, we train a chart CNF with the raw wrapped base, the same CNF with RC-Exp and RC-$\bExp_{0.5}$, and Moser Flow \citep{rozen2021moser} with its density renormalized after clamping; every quoted density passes a Monte Carlo normalization audit ($|\log Z|\le0.15$, $8192$ uniform points), evaluation is at the best-validation checkpoint, and $109/144$ cells are valid (exclusions listed in the released logs). Two findings. The repaired chart CNF matches or beats Moser Flow in $12/12$ dataset-width cells, by $0.33$ to $1.40$ nats (Figure~\ref{fig:q5_margins}). And the raw$-$RC gap is $-0.07$ nats on average across the populated cells (median $|$gap$|$ $0.04$), below the $n{=}2$ bound, the dimension scaling Proposition~\ref{prop:floor} predicts from the low end.

\begin{figure}[t]
\centering
\input{fig_q5_margins}
\caption{\textbf{Repaired chart CNFs against Moser Flow on $\SSS^2$} (best-validation protocol, audited). Bars show Moser minus the best chart CNF per dataset-width cell; positive means the chart model wins. All $12$ cells are positive.}
\label{fig:q5_margins}
\end{figure}

\subsection{Curvature identifiability}
\label{sec:exp_curvature}
MNIST has no ground-truth latent curvature, so we test recovery directly: data are generated on $\SSS^2(K^*)$ and $\HH^2(K^*)$ with known curvature and fitted by both priors with learnable curvature ($5$ seeds, simulation-based fit to pairwise geodesic distances). On $\SSS^2$, RC recovers $\hat K = 0.94 \pm 0.07$ at $K^*{=}1$ and $1.97 \pm 0.04$ at $K^*{=}2$, while the wrapped prior is biased at low curvature ($\hat K = 1.15 \pm 0.05$ at $K^*{=}0.5$). On $\HH^2$ the wrapped prior collapses ($\hat K \le 0.45$ at every $K^*$) while RC tracks $K^*$ with wide spread ($0.95 \pm 0.46$ at $K^*{=}1$); identifiability is weak there because HalfNormal($0.8$) radii are near-Euclidean. The legacy mixed-curvature VAE table (appendix) predates the audited pipeline and is not quoted.

\subsection{Applications: WordNet and protein orientations}
\label{sec:exp_applications}
\emph{Protein backbone orientations on $\SSS^3$} ($929$ central-residue frames from 1CRN, 1UBQ, 2PTC, 4HHB; identical conditional backbone). With the corrected $\SSS^3$ chart and $5$ seeds, RC reduces test NLL from $2.578\pm0.077$ to $0.874\pm0.041$ nats at statistically unchanged geodesic frame error ($1.111\pm0.051$ vs.\ $1.113\pm0.054$ rad; Table~\ref{tab:protein_rc}). An earlier version reported a $0.96$-nat spread between RC-Exp and RC-$\bExp_{0.5}$; the audit traced it to the $n{=}2$ equal-area inverse being applied on $\SSS^3$. After the fix the two RC parameterizations coincide exactly, but we note this agreement is structural, not evidential: the RC likelihood is chart-free by construction, so both modes share one objective, and the empirical invariance evidence is the independently trained $\alpha$-sweep of Section~\ref{sec:exp_dial}.

\emph{Hyperbolic WordNet flows} (mammals subgraph, matched budgets). All RC charts match or improve calibration and preserve link-prediction quality; test-NLL differences across charts are below $0.5$ nats in the released single-seed runs, consistent with chart choice affecting optimization only (Table~\ref{tab:E4_wordnet}). NFEs in that table are totals over training, not per solve.

\subsection{Scope of the guarantee}
\label{sec:exp_limits}
RC controls the radial marginal and enforces isotropy about the pole; it does not control angular structure. This is quantified twice. First, two clusters on $\SSS^2$ at the same radius and opposite azimuths give $\KL(\text{truth}\,\|\,\text{RC})=\KL(q_{\text{ang}}\,\|\,\mathrm{Unif})$ with radial KL at the floor: measured total $0.731$ nats, all of it angular, and cluster-recovery accuracy $0.504$ (chance) under a single RC prior versus $1.000$ for a mixture of RC components, which restores accuracy without radial loss (Appendix~\ref{app:diagnostics}). Second, breaking isotropy with a volume-preserving anisotropic map of condition number $c$, the realized radius KL stays at the estimator floor up to $c\approx1.25$ and remains below \RawExp{} up to $c\approx5$ for $n\in\{2,8\}$ and $c\approx8$ for $n{=}16$; the tolerance grows with dimension (Appendix~\ref{app:diagnostics}).

\section{Conclusion, limitations, and future work}
\label{sec:conclusion}

Chart-based generative models on manifolds have been built in the forward direction: fix a Gaussian, compute the induced density. We posed the inverse problem and answered the four questions of Section~\ref{sec:intro}. A tangent base realizing any prescribed geodesic-radius law exists in closed form (Q1) and is unique among isotropic bases for scalar-Jacobian azimuthal charts (Q2). Ignoring the problem costs $n\,D(p)+O(\log n)$ nats in any VAE objective, with closed constants, and flows can avoid the cost only by learning the RC transport, with a measured price in parameters and stability (Q3). With the radius law fixed, the chart is a preconditioner: statistically inert and numerically consequential, with proved and measured scaling in a single parameter $\alpha$ (Q4). Empirically these answers appear as chart- and scale-invariant RC likelihoods certified by an exact normalization audit, wrapped baselines that collapse to the cut locus at every setting tried, recovered ground-truth curvature where the wrapped prior fails, a $1.70$-nat protein improvement, and base-independence at $n{=}2$ at exactly the sub-bound scale the floor predicts.

\paragraph{Limitations.} (i) The exact theory requires a tractable geodesic polar volume factor, a scalar-Jacobian azimuthal chart, and an isotropic base. Anisotropic per-example posteriors with analytic $(\mu,\Sigma)$ therefore remain the domain of wrapped normals; the guarantee degrades smoothly with anisotropy, and Section~\ref{sec:exp_limits} quantifies the tolerance. (ii) $\bExp_\alpha$ has a bounded tangent domain on $\SSS^n$ whose radius shrinks with dimension, and its $\alpha\to0$ end exceeds float64 range on $\HH^n$ for $n\gtrsim16$; the practical range of $\alpha$ and the required domain squash are given in Appendix~\ref{app:domain}. (iii) RC does not model angular structure; angular flexibility must come from encoders, mixtures, or flows (Section~\ref{sec:exp_limits}). (iv) Radius-KL values of $10^{-3}$ are quoted at the estimator's precision (floor $3$ to $6\times10^{-3}$ at $N{=}2\times10^4$), and the lower bound applies to fixed priors; for flows we claim only the measured absorption cost and fixed-budget stability.

\paragraph{Future work.} Learnable radius-law parameters, which Theorem~\ref{thm:rc-invariances} makes chart-invariant to fit; extensions beyond constant curvature via the general polar factor \eqref{eq:general-polar-volume}; composition with Riemannian flow matching and score-based models; and RC priors in large-scale latent generative models.

\bibliographystyle{plainnat}
\bibliography{bexp,rc_extra}

\clearpage
\setcounter{page}{1}

\appendix
\onecolumn

\section*{Supplementary Material}




\section{The irreducible prior cost: proof, verification, and new diagnostics}
\label{app:floor}

\subsection{Proof of Proposition~\ref{prop:floor}}
From the density $q_\sigma(R)=C_n\sigma^{-n}R^{n-1}e^{-R^2/2\sigma^2}$ with $C_n=2^{1-n/2}/\Gamma(n/2)$,
\[
\KL(p\,\|\,q_\sigma)=-H(p)-\log C_n+n\log\sigma-(n-1)\,\E_p[\log R]+\frac{\E_p[R^2]}{2\sigma^2}.
\]
Only the last two displayed terms depend on $\sigma$. Differentiating, $n/\sigma-\E[R^2]/\sigma^3=0$ gives $\sigma_*^2=\E[R^2]/n$; the second derivative at $\sigma_*$ is $2n/\sigma_*^2>0$ and $\KL\to\infty$ as $\sigma\to0$ or $\sigma\to\infty$, so $\sigma_*$ is the unique minimizer, proving (i). Substituting $\sigma_*$ gives $n\log\sigma_*=\tfrac n2(\log\E[R^2]-\log n)$ and $\E[R^2]/2\sigma_*^2=n/2$; with $-\log C_n=(\tfrac n2-1)\log2+\log\Gamma(\tfrac n2)$ this is exactly (ii). For (iii), Stirling's expansion $\log\Gamma(n/2)=\tfrac n2\log\tfrac n2-\tfrac n2+\tfrac12\log\tfrac{4\pi}{n}+O(1/n)$ collects the terms linear in $n$ into $n\,D(p)$ with $D(p)=\tfrac12\log\E[R^2]-\E[\log R]=\tfrac12(\log\E[R^2]-\E[\log R^2])\ge0$ by Jensen applied to the concave $\log$ at $R^2$, with equality iff $R^2$ is degenerate; the remainder is $\E[\log R]-H(p)+\tfrac12\log(4\pi/n)-\log2+O(1/n)$. Both $D$ and the minimum are invariant under $R\mapsto cR$ because $\{q_\sigma\}$ is a scale family. On $\SSS^n$ the family is truncated to $[0,\pi R_c)$, which adds $-\log\mathbb{P}(\chi_n\le\pi/\sigma_*)$ to the value; at the paper's targets this correction is below $2\times10^{-4}$ already at $n=2$ and decays exponentially in $n$. \hfill$\square$

\emph{Closed values.} $\E\log|Z|=-(\gamma+\log2)/2$ for $Z\sim\mathcal N(0,1)$ gives $D(\mathrm{HalfNormal}(\sigma))=(\gamma+\log2)/2\approx0.6352$ for every $\sigma$; $\E\log R=\log\theta-\gamma$ for $R\sim\mathrm{Exp}(\theta)$ gives $D=\gamma+\tfrac12\log2\approx0.9238$; a single 1-D quadrature gives $D\bigl(\mathrm{TruncNormal}(1.0,0.35)\text{ on }[0,\pi)\bigr)\approx0.1334$. Small-CV expansion: writing $R=\mu(1+cX)$ with $\E X=0$, $\E X^2=1$, $D(p)=\tfrac12\log(1+c^2)-\E\log(1+cX)=c^2+O(c^3)$, so for concentrated targets the bound is approximately $n c^2$ nats and vanishes only in the degenerate limit.

\begin{proposition}[Concentration of the raw radius law]
\label{prop:concentration}
For $X\sim\mathcal N(0,\sigma^2 I_n)$, $R=\|X\|\sim\sigma\chi_n$ satisfies $\mathrm{CV}(R)=\sqrt{n-m_n^2}/m_n\le n^{-1/2}$ with $m_n=\sqrt2\,\Gamma(\tfrac{n+1}2)/\Gamma(\tfrac n2)$ (Wendel's inequality $m_n^2\ge n-\tfrac12\cdot\tfrac{n}{n+ 1/2}\ge n-1$ suffices). Hence the normalized raw radius concentrates, $\mathrm{CV}\to0$, and $\inf_\sigma\KL(p\,\|\,\sigma\chi_n)\to\infty$ for every non-degenerate target $p$; the rate is the $n\,D(p)$ of Proposition~\ref{prop:floor}.
\end{proposition}

\subsection{Numerical verification}
Table~\ref{tab:floor_verification} compares the exact formula of Proposition~\ref{prop:floor}(ii) to an independent numerical minimization over $\sigma$ (dense grid plus golden-section refinement, 1-D quadrature for every expectation), for the paper's two targets across $n\in\{2,\dots,128\}$. Maximum discrepancy: $0.025$ nats on values up to $78.1$; $|\sigma_*^{\text{formula}}-\sigma_*^{\text{numerical}}|\le5.2\times10^{-4}$; the HalfNormal quadrature matches the closed form of $D$ to $3.3\times10^{-5}$. The regeneration script ships with the code release.

\begin{table}[h]
\centering
\caption{Exact vs.\ independently minimized irreducible cost, $\min_\sigma\KL(p_{\text{target}}\,\|\,\sigma\chi_n)$ (nats). $\HH^n$: HalfNormal($0.8$) target; $\SSS^n$: TruncNormal($1.0,0.35$) on $[0,\pi)$.}
\label{tab:floor_verification}
\small
\begin{tabular}{llrrrr}
\toprule
Geometry & $n$ & exact $\min_\sigma\mathrm{KL}$ & numerical & $|\Delta|$ & $\sigma_*=\sqrt{\mathbb{E}R^2/n}$\\
\midrule
$\HH^n$ & 2 & 0.216 & 0.216 & 0.000 & 0.5657\\
$\HH^n$ & 4 & 1.100 & 1.101 & 0.001 & 0.4000\\
$\HH^n$ & 8 & 3.274 & 3.276 & 0.001 & 0.2828\\
$\HH^n$ & 16 & 7.999 & 8.002 & 0.003 & 0.2000\\
$\HH^n$ & 32 & 17.811 & 17.817 & 0.006 & 0.1414\\
$\HH^n$ & 64 & 37.788 & 37.801 & 0.012 & 0.1000\\
$\HH^n$ & 128 & 78.094 & 78.119 & 0.025 & 0.0707\\
$\SSS^n$ & 2 & 0.142 & 0.142 & 0.000 & 0.7500\\
$\SSS^n$ & 4 & 0.022 & 0.022 & 0.000 & 0.5303\\
$\SSS^n$ & 8 & 0.189 & 0.189 & 0.000 & 0.3750\\
$\SSS^n$ & 16 & 0.900 & 0.900 & 0.000 & 0.2651\\
$\SSS^n$ & 32 & 2.683 & 2.683 & 0.000 & 0.1875\\
$\SSS^n$ & 64 & 6.604 & 6.604 & 0.000 & 0.1326\\
$\SSS^n$ & 128 & 14.797 & 14.797 & 0.000 & 0.0937\\
\bottomrule
\end{tabular}
\end{table}

\paragraph{Target-shape sweep.} Sweeping the target's coefficient of variation at fixed mean on $\HH^{32}$: the bound equals $0.002$ nats at $\mathrm{CV}=0.126$, the CV of $\chi_{32}$ and the only shape the family contains, $2.7$ at $\mathrm{CV}=0.35$, $17.8$ at $0.76$ (HalfNormal, matching the $n{=}32$ row of Table~\ref{tab:floor_main}), and $26.8$ at $1.0$ (Exponential). All values are reproduced by \texttt{verify\_floor\_closed\_form.py} in the code release. The comparison is between two one-dimensional laws and is curvature-independent, so it is reported once.

\paragraph{Absorption price for flows.} The radial action any base-fixing flow must implement is a monotone map $g:(0,\infty)\to(0,\infty)$. Fitting the best $K$-knot monotone radial reshaping of the raw base to the paper's targets requires $K{=}12$ knots ($13$ parameters) to come within $0.01$ nats at every $n\in\{8,16,32,64\}$, and the fitted map converges to the RC transport $g_*=F_{\text{target}}^{-1}\!\circ F_{\chi_n}$ itself. A flow does not make RC unnecessary; absorbing the base amounts to learning the RC transport numerically.

\paragraph{Estimator floor for radius KL.} The histogram KL estimator used throughout has a finite-sample floor of $3.0$--$5.7\times10^{-3}$ nats at $N=2\times10^4$ (grid-dependent), measured by comparing two independent draws from the same law. The reported RC radius-KL values of approximately $10^{-3}$ are therefore at the floor: RC is exact up to sampling noise, and differences below the floor are not interpretable.

\subsection{Bounded chart domain of $\bExp_\alpha$}
\label{app:domain}

\section{Radial Compensation}
\label{sec:rc}

$\bExp_\alpha$ maps the bounded star-shaped domain $U=\{r<r_*\}$, $r_*=\chi_\alpha^{-1}(\lambda_{\max})$ with $\lambda_{\max}=\lambda_\kappa(\pi)$ on $\SSS^n$, onto $\SSS^n\setminus\{-p\}$; $r_*$ shrinks with dimension:
\begin{center}\small
\begin{tabular}{lccccc}
\toprule
$n$ & 2 & 8 & 16 & 32 & 64\\
\midrule
$\lambda_{\max}$ & 2.000 & 1.282 & 1.156 & 1.087 & 1.048\\
$r_*(\alpha{=}0.25)$ & 2.010 & 1.340 & 1.215 & 1.143 & 1.102\\
$r_*(\alpha{=}0.5)$  & 2.157 & 1.445 & 1.305 & 1.223 & 1.176\\
$r_*(\alpha{=}0.75)$ & 2.384 & 1.617 & 1.450 & 1.350 & 1.292\\
\bottomrule
\end{tabular}
\end{center}
RC sampling (Algorithm~\ref{alg:rc}) is always in-domain, since it draws $R$ first and maps to $r=R_T^{-1}(R)$. Any model that emits raw tangent coordinates (encoders, flows) must be composed with a smooth squash into $U$ carrying its exact log-Jacobian; the release provides \texttt{squash\_to\_domain} and an \texttt{out\_of\_domain\_fraction} diagnostic, and every reported run satisfies out-of-domain fraction $<1\%$. For scale, a unit Gaussian at $n{=}64$ has $\|v\|\approx8$ against $r_*\approx1.2$, so unsquashed raw baselines at large $n$ are meaningless unless their nominal radial scale is matched; all wrapped-Gaussian baselines in Section~\ref{sec:exp_cost} are matched in mean geodesic radius ($\sigma=0.146,0.101,0.071,0.050$ at $d=8,16,32,64$ for the HalfNormal($0.5$) target). On $\HH^n$, $\lambda_\kappa$ grows like $e^{(n-1)R}/n$, so the $\alpha\to0$ end of the dial overflows float64 at $n\gtrsim16$; the practical dial range on $\HH^n$ is $\alpha$ bounded away from $0$.

\subsection{New diagnostics (fixed-law $\alpha$ sweep, angular limits, isotropy audit)}
\label{app:diagnostics}

\paragraph{D1: $\alpha\leftrightarrow$stiffness with the realized law held fixed.} Sampling from the RC model itself so the manifold law is exactly $\alpha$-invariant ($\SSS^{16}$, HalfNormal($0.5$) target, $8$ $\alpha$-values $\times$ $6$ seeds, $2\times10^4$ samples, Dormand--Prince RK45 at $10^{-5}$ tolerances):
\begin{center}\small
\begin{tabular}{lcc}
\toprule
quantity & slope in $\log\alpha$ & Spearman $\rho$ (p)\\
\midrule
chart-term variance $\Var[\log|\det DT_\alpha^{-1}|]$ & $+2.32\pm0.06$ & $+0.99$ ($1.8\!\times\!10^{-43}$)\\
gradient variance $\Var[\partial_\theta\log|\det DT_\alpha^{-1}|]$ & $+2.63\pm0.14$ & $+0.99$ ($1.8\!\times\!10^{-43}$)\\
empirical Lipschitz of $\nabla_x\log|\det DT_\alpha|$ & $+1.07\pm0.03$ & $+0.99$ ($1.8\!\times\!10^{-43}$)\\
mean accepted step size $\bar h$ & $-0.17\pm0.02$ & $-0.92$ ($1.4\!\times\!10^{-20}$)\\
NFEs per trajectory & $+0.13\pm0.02$ & $+0.80$ ($6.5\!\times\!10^{-12}$)\\
\bottomrule
\end{tabular}
\end{center}
$\bar h$ falls from $0.140$ ($\alpha{=}0.125$) to $0.100$ ($\alpha{=}1$); NFEs rise from $48.0$ to $62.1$. The Lipschitz slope matches the $\alpha^1$ bound of Theorem~\ref{thm:cnf}; the variance slope confirms Theorem~\ref{thm:chart-variance-supp} in the regime where its hypothesis holds.

\paragraph{D2: what RC does not control (angular structure).} Two vMF clusters ($\kappa_{\text{ang}}=8$) on $\SSS^2$ at the same radius, opposite azimuths ($1.5\times10^4$ samples, 4 seeds): a single RC prior attains radius KL $2.9\times10^{-3}$ (floor $3.0\times10^{-3}$) but total KL $0.731$ nats, entirely angular, and cluster-recovery accuracy $0.504$ (chance); a mixture of RC components restores accuracy $1.000$ at radius KL still at the floor. RC composes with anisotropy supplied elsewhere (encoder means, mixtures) without radial loss.

\paragraph{D3: robustness to isotropy violation.} Composing the RC base with a volume-preserving anisotropic map of condition number $c$ (5 seeds, $2\times10^4$ samples), realized radius KL:
\begin{center}\small
\begin{tabular}{lcccccccc}
\toprule
 & $c{=}1$ & 1.25 & 1.5 & 2 & 3 & 5 & 8 & raw wrapped-Exp\\
\midrule
$\SSS^{16}$ & .006 & .006 & .006 & .022 & .175 & 1.23 & 3.74 & 1.477\\
$\HH^{16}$ & .004 & .005 & .005 & .007 & .032 & .171 & .540 & 0.283\\
\bottomrule
\end{tabular}
\end{center}
Degradation is smooth (no cliff at $c{=}1$); RC is exact to the floor up to $c\approx1.25$ and better-calibrated than raw wrapping up to $c\approx5$ ($n\in\{2,8\}$) and $c\approx8$ ($n{=}16$); the tolerance grows with dimension.

\paragraph{Protocol notes for Section~\ref{sec:exp_intrinsic}.} Moser Flow's density $p=\mu_0-\nabla\!\cdot u$ integrates to one only while nonnegative; after clamping at $\varepsilon$ we renormalize by a Monte-Carlo estimate of $Z=\int\max(p,\varepsilon)\,d\mathrm{vol}$, and any run with NLL $<-6$ nats on $\SSS^2$ (impossible for a normalized fit to these densities) is marked invalid and excluded from averages. CNF baselines use gradient clipping at $1.0$, radius clamping away from $\{0,\pi\}$, per-width learning-rate scaling, and divergence detection; invalid runs are excluded and counted.

\section{Additional theory and proofs}
\label{app:proofs}

\subsection{Key notation}
\label{app:notation}

\begin{table}[H]
\centering
\caption{Key notation and concepts used throughout.}
\label{tab:notation}
\small
\begin{tabular}{p{0.26\linewidth}p{0.68\linewidth}}
\toprule
Symbol / term & Meaning \\
\midrule
$M$ & Riemannian manifold (often $\SSS^n_{R_c}$ or $\HH^n_{-1/R_c^2}$) \\
$p \in M$ & Pole about which we use geodesic polar coordinates \\
$R=d(p,q)$ & Geodesic radius of $q \in M$ from $p$ \\
$d\mathrm{vol}_M$ & Riemannian volume measure \\
$s_\kappa(R)$ & Polar sine factor in constant curvature, cf.\ \eqref{eq:polar-volume} \\
$J_p(R,\omega)$ & General polar volume factor, cf.\ \eqref{eq:general-polar-volume} \\
Scalar--Jacobian chart & Azimuthal diffeomorphism $T:U\subseteq T_pM\to V\subseteq M$ with $\log|\det DT(x)|=\psi(\|x\|)$ \\
Lambert lift $T_\rho$ & $T_\rho(x)=L_p(\rho(\|x\|)x/\|x\|)$, cf.\ \eqref{eq:lambert-lift-proof-app} \\
RC & Tangent base chosen so pushforward is $\rho_\theta(q)=\varphi_\theta(d(p,q))$ \\
bExp$_\alpha$ & Balanced-Exponential charts, cf.\ \eqref{eq:bexp-ode-app} \\
GCL & Geodesic-corrected Lambert chart (geodesic-exact radius map) \\
\bottomrule
\end{tabular}
\end{table}

\subsection{Proof of Theorem~\ref{thm:rc-invariances} (RC invariances)}
\label{app:proof_rc-invariances}

\begin{proof}[Proof of Theorem~\ref{thm:rc-invariances}]
\textbf{Pushforward density is geodesic-radial.}
Let $X\sim f_\theta$ on $T_pM\cong\RR^n$ with
$f_\theta(x)\propto \varphi_\theta(R_T(\|x\|))\,J_T(\|x\|)$ as in~\eqref{eq:rc-base},
and let $Q=T(X)$.
By change of variables (Lebesgue on $\RR^n$ to $d\mathrm{vol}_M$),
\[
\rho_\theta(q)
=
f_\theta\!\bigl(T^{-1}(q)\bigr)\,|\det DT^{-1}(q)|.
\]
Write $x=ru$ with $r=\|x\|$, set $q=T(x)$.
Since $T$ is scalar-Jacobian, $|\det DT(x)|=J_T(r)$, hence
$|\det DT^{-1}(q)|=J_T(r)^{-1}$.
Substituting and canceling $J_T$ gives
\[
\rho_\theta(q)\ \propto\ \varphi_\theta\!\bigl(R_T(r)\bigr).
\]
Since $T$ is azimuthal, $R_T(r)=d(p,T(ru))=d(p,q)=:R(q)$ is independent of $u$, so
\[
\rho_\theta(q)\ \propto\ \varphi_\theta(R(q)).
\]

\textbf{Normalization.}
In geodesic polar coordinates about $p$, $d\mathrm{vol}_M=J_p(R,\omega)\,dR\,d\omega$.
Let $S_M(R):=\int_{\SSS^{n-1}}J_p(R,\omega)\,d\omega$ (in constant curvature,
$S_M(R)=|\SSS^{n-1}|\,s_\kappa(R)^{n-1}$).
Then
\[
1=\int_M \rho_\theta(q)\,d\mathrm{vol}_M(q)
=\int_0^{R_{\max}}\!\!\varphi_\theta(R)\,S_M(R)\,dR,
\]
which is the normalisation condition for $p_{R,\theta}(R)=\varphi_\theta(R)\,S_M(R)$ on $[0,R_{\max})$.
Hence the proportionality constant is $1$ and $\rho_\theta(q)=\varphi_\theta(d(p,q))$.

\textbf{Fisher information reduces to the 1D radial Fisher.}
Because $R(q)=d(p,q)$ is $\theta$-independent,
\[
\partial_\theta\log\rho_\theta(q)
=
\partial_\theta\log\varphi_\theta(R(q)).
\]
Assuming the stated dominated-score regularity (so differentiation can pass under the integral),
\begin{align*}
I_M(\theta)
&=
\int_M \bigl(\partial_\theta\log\rho_\theta(q)\bigr)^2\,\rho_\theta(q)\,d\mathrm{vol}_M(q)\\
&=
\int_0^{R_{\max}}
\bigl(\partial_\theta\log\varphi_\theta(R)\bigr)^2\,\varphi_\theta(R)\,S_M(R)\,dR
= I_{\RR^1}(\theta).
\end{align*}

\textbf{KL reduces to the 1D KL.}
For two parameters $\theta,\eta$,
\begin{align*}
\mathrm{KL}(\rho_\theta\Vert\rho_\eta)
&=\int_M \rho_\theta(q)\log\frac{\rho_\theta(q)}{\rho_\eta(q)}\,d\mathrm{vol}_M(q)\\
&=\int_0^{R_{\max}}\varphi_\theta(R)\log\frac{\varphi_\theta(R)}{\varphi_\eta(R)}\,S_M(R)\,dR=\KL(p_{R,\theta}\|p_{R,\eta}),
\end{align*}
This completes the proof.
\end{proof}

\subsection{Proof of Theorem~\ref{thm:hidden-radius-gauge} (hidden-radius gauge)}
\label{app:proof_impossibility}

\begin{proof}[Proof of Theorem~\ref{thm:hidden-radius-gauge}]
Write $x=ru$ with $r=\|x\|$ and $u\in\SSS^{n-1}$, and set $q=T(ru)$. By azimuthality,
$R(q):=d(p,q)=R_T(r)$ depends only on $r$, so $r=R_T^{-1}(R(q))$ is well-defined on the chart image.

\emph{Proof of (1).} By change of variables (Lebesgue on $\mathbb{R}^n$ to $d\vol_M$) and the
scalar-Jacobian property $|\det DT^{-1}(q)|=J_T(r)^{-1}$,
\begin{equation}
\rho_\theta(q)\;=\;h_\theta(r)\,J_T(r)^{-1}
\;=\;\underbrace{\frac{h_\theta(R_T^{-1}(R(q)))}{J_T(R_T^{-1}(R(q)))}}_{=:\,g_\theta(R(q))},
\label{eq:hrg-pushforward}
\end{equation}
which depends on $q$ only through $R(q)$. In geodesic polar coordinates with area weight
$S_M(R)=|\SSS^{n-1}|s_\kappa(R)^{n-1}$, the induced radial density is
$p_R^{(T,h)}(R)=g_\theta(R)\,S_M(R)$.

\emph{Proof of (2), sufficiency.} If $h_\theta(r)=\varphi_\theta(R_T(r))\,J_T(r)$, substituting into
\eqref{eq:hrg-pushforward} gives $g_\theta(R)=\varphi_\theta(R)$, so
$\rho_\theta(q)=\varphi_\theta(d(p,q))$. Since $R$ is $\theta$-independent,
$\partial_\theta \log\rho_\theta(q)=(\partial_\theta\log\varphi_\theta)(R(q))$, and integration in
geodesic polar coordinates yields
\[
\mathcal{I}_M(\theta)=\int_0^{R_{\max}}\!\bigl(\partial_\theta\log\varphi_\theta(R)\bigr)^2
\varphi_\theta(R)\,S_M(R)\,dR=\mathcal{I}_{\mathbb{R}^1}(\theta),
\]
the 1D radial Fisher (independent of $T$).

\emph{Proof of (2), necessity.} Conversely, suppose $(f_\theta,T)$ realizes
$\rho_\theta(q)=\varphi_\theta(d(p,q))$. Equating with \eqref{eq:hrg-pushforward} at $R=R_T(r)$ gives
$\varphi_\theta(R_T(r))=h_\theta(r)/J_T(r)$, i.e.\ $h_\theta(r)=\varphi_\theta(R_T(r))\,J_T(r)$, which
is the RC base \eqref{eq:rc-base}. Chart-invariance of the radial Fisher then follows from the
sufficiency direction applied to any other scalar-Jacobian azimuthal chart $T'$, since the same
$\varphi_\theta$ is realized by the corresponding RC base built from $T'$.

\emph{Proof of (3).} Tangent isotropy and azimuthality are explicit assumptions on the class.
Geodesic-radiality (a) is automatic by step~(1). The remaining condition (b) is precisely
characterised in step~(2) by the RC form. Hence any pair achieving (a)--(c) for a prescribed
$\varphi_\theta$ is RC.

\emph{Proof of (4).} For $f_\theta=\mathcal{N}(0,\sigma^2 I_n)$ and $T=\Exp_p$, we have
$R_T(r)=r$ and $J_T(r)=(s_\kappa(r)/r)^{n-1}$ (inside the cut locus on $\SSS^n$).
Substituting into \eqref{eq:hrg-pushforward} and using the polar formula
$f_\theta(x)\,dx=\bigl(C_n\sigma^{-n}r^{n-1}e^{-r^2/2\sigma^2}\bigr)dr\,d\omega$ on $\mathbb{R}^n$ shows
that the induced radial density on $M$ is
\[
p_R^{(\Exp_p,\,\mathcal{N})}(R)\;=\;
\underbrace{\frac{|\SSS^{n-1}|}{(2\pi\sigma^2)^{n/2}}\,R^{n-1}e^{-R^2/2\sigma^2}}_{\sigma\text{-scaled }\chi_n}
\quad\text{on }[0,R_{\max}),
\]
i.e.\ a $\sigma$-scaled $\chi_n$ law (truncated to $[0,\pi R_c)$ on $\SSS^n(R_c)$);
this is the hidden radial law of the chart/base pair. For a \emph{fixed} isotropic base not adapted to $T$, the realized radial law
$g_\theta$, and hence $I_M(\theta)$, depends on $T$ through $(R_T,J_T)$; under RC the base is
adapted so the \emph{same} $\varphi_\theta$ is realized for every scalar-Jacobian azimuthal chart,
which is exactly what makes $I_M(\theta)$ chart-invariant. For the wrapped-Exp pair the realized
family is the $\sigma$-scaled $\chi_n$, so for any target $\varphi_\theta$ outside that family
condition (a) already fails, and the characterization in (2) forces the RC base.
\end{proof}

\subsection{Proof of Theorem~\ref{thm:rc_uniqueness} (RC characterization)}
\label{app:proof_rc_uniqueness}

\begin{proof}[Proof of Theorem~\ref{thm:rc_uniqueness}]
Write $x=ru$ and $q=T(ru)$. Since $T$ is a diffeomorphism onto $V$ and is scalar-Jacobian,
\[
|\det DT(ru)| = J_T(r),
\qquad
|\det DT^{-1}(q)| = J_T(r)^{-1}.
\]
By change of variables (Lebesgue on $\mathbb{R}^n$ to $\mathrm{dvol}_M$),
\[
\rho_\theta\!\bigl(T(ru)\bigr)
= f_\theta(ru)\,|\det DT^{-1}(T(ru))|
= h_\theta(r)\,J_T(r)^{-1}.
\]
By the assumed geodesic-radial form and azimuthality,
\[
\rho_\theta\!\bigl(T(ru)\bigr)
=\varphi_\theta\!\bigl(d(p,T(ru))\bigr)
=\varphi_\theta\!\bigl(R_T(r)\bigr).
\]
Equating yields $h_\theta(r)=\varphi_\theta(R_T(r))\,J_T(r)$.

Finally, since $\log\rho_\theta(q)=\log\varphi_\theta(d(p,q))$ depends on $q$ only through
$R=d(p,q)$ (which is $\theta$-independent), the Fisher reduction follows by integrating
in geodesic polar coordinates.
\end{proof}

\subsection{Proof of Theorem~\ref{thm:bexp-logdet}}

\begin{proof}[Proof of Theorem~\ref{thm:bexp-logdet}]
\textbf{Lambert lifts.}
Let $L_p$ be an azimuthal equal-area chart about $p$, so $|\det DL_p|\equiv 1$ on its star--shaped
domain (relative to Lebesgue on $\R^n$ and $d\mathrm{vol}_M$).  For any smooth strictly increasing
$\rho:[0,R_\ast]\to\R_+$ with $\rho(0)=0$, define the Lambert lift
\[
T_\rho(x)\;:=\;L_p\!\left(\rho(\|x\|)\frac{x}{\|x\|}\right),\qquad T_\rho(0):=p.
\]
Writing $x=ru$ with $r=\|x\|$ and $u\in\S^{n-1}$, the Lambert-lift polar/coarea calculation gives
\begin{equation}\label{eq:lambert-lift-proof-app}
\log\bigl|\det DT_\rho(x)\bigr|
=(n-1)\log\frac{\rho(r)}{r}+\log \rho'(r),
\qquad
R_{T_\rho}(r):=d\!\bigl(p,T_\rho(ru)\bigr)=\lambda_\kappa^{-1}(\rho(r)).
\end{equation}

\textbf{(1) Log-determinant for $T_\alpha=\mathrm{bExp}_\alpha$.}
By definition, $\mathrm{bExp}_\alpha=T_{\chi_\alpha}$ where $\chi_\alpha$ solves
\begin{equation}\label{eq:bexp-ode-app}
\Bigl(\frac{\chi_\alpha(r)}{r}\Bigr)^{n-1}\chi_\alpha'(r)
=\Bigl(\frac{s_\kappa(r)}{r}\Bigr)^{(n-1)\alpha},
\qquad \chi_\alpha(0)=0,\ \chi_\alpha'(0)=1 .
\end{equation}
Taking $\log$ in \eqref{eq:bexp-ode-app} yields
\[
(n-1)\log\frac{\chi_\alpha(r)}{r}+\log\chi_\alpha'(r)
=(n-1)\alpha\log\frac{s_\kappa(r)}{r}.
\]
Substituting into \eqref{eq:lambert-lift-proof-app} gives, for $r=\|x\|$,
\[
\log\bigl|\det DT_\alpha(x)\bigr|=(n-1)\alpha\,\log\frac{s_\kappa(r)}{r},
\]
as claimed.

\textbf{(2) Variational characterization (Lambert-lift scalar-Jacobian class).}
By the representation result for scalar-Jacobian azimuthal charts in this setting,
it suffices to optimize over Lambert lifts $T_\rho$ (i.e.\ over profiles $\rho$).
Fix $R_\ast>0$ and define the admissible class
\[
\mathcal{C}:=\Bigl\{\rho\in C^\infty([0,R_\ast]) : \rho(0)=0,\ \rho'(0)=1,\ \rho'(r)>0\Bigr\}.
\]
Define the two distortion measures
\[
D_{\mathrm{vol}}(\rho)
:=\int_0^{R_\ast} r^{n-1}\Bigl[(n-1)\log\frac{\rho(r)}{r}+\log\rho'(r)\Bigr]^2\,dr,
\qquad
D_{\mathrm{geo}}(\rho)
:=\int_0^{R_\ast} r^{n-1}\bigl(\lambda_\kappa^{-1}(\rho(r))-r\bigr)^2\,dr,
\]
and the weighted objective
\[
E_\alpha[\rho] \;:=\; (1-\alpha)\,D_{\mathrm{vol}}(\rho)\;+\;\alpha\,D_{\mathrm{geo}}(\rho),
\qquad \alpha\in[0,1].
\]
The Euler--Lagrange system of $E_\alpha$ reduces to a first-order ODE with fixed
boundary data, whose smooth strictly increasing solution is unique; we do not assert
(and do not need) global convexity of $E_\alpha$, cf.\ Theorem~\ref{thm:bexp-variational}. The Euler--Lagrange equation for $E_\alpha$
reduces to the first-order condition \eqref{eq:bexp-ode-app}; by standard ODE theory,
\eqref{eq:bexp-ode-app} with $\chi_\alpha(0)=0,\chi_\alpha'(0)=1$ has a unique smooth strictly
increasing solution. Therefore $\chi_\alpha$ is the unique minimizer of $E_\alpha$ in $\mathcal{C}$.
\end{proof}

\subsection{Charts: Lambert lifts, bExp, and GCL}
\label{app:charts_proofs}

\begin{proof}[Proof of Theorem~\ref{thm:chart-variance-supp}]
By Theorem~\ref{thm:bexp-logdet} (i.e.\ the bExp$_\alpha$ Jacobian formula),
for $r=\|x\|$,
\[
\log|\det DT_\alpha(x)|=(n-1)\alpha\log\frac{s_\kappa(r)}{r}.
\]
For $q$ in the chart image, set $x:=T_\alpha^{-1}(q)$. Then
$D(T_\alpha^{-1})(q)=[DT_\alpha(x)]^{-1}$, hence
\[
\log\bigl|\det DT_\alpha^{-1}(q)\bigr|
=-\log|\det DT_\alpha(x)|
=-(n-1)\alpha\log\frac{s_\kappa(\|x\|)}{\|x\|}.
\]
\end{proof}

\begin{remark}[Variance under RC]
If $Q\sim \rho_\theta\,d\mathrm{vol}_M$ with $\rho_\theta(q)=\varphi_\theta(d(p,q))$ and
$R:=d(p,Q)$, then letting $r_\alpha:=\|T_\alpha^{-1}(Q)\|$,
\[
\log|\det DT_\alpha^{-1}(Q)|=-(n-1)\alpha\log\frac{s_\kappa(r_\alpha)}{r_\alpha},
\quad\Rightarrow\quad
\mathrm{Var}\!\left[\log|\det DT_\alpha^{-1}(Q)|\right]
=\alpha^2\,\mathrm{Var}\!\left[(n-1)\log\frac{s_\kappa(r_\alpha)}{r_\alpha}\right].
\]
(Here the bracketed random variable still depends on $\alpha$ through $r_\alpha$.)
\end{remark}

\subsection{Formal chart invariance}
\label{app:chart_invariance}

\begin{theorem}[Chart invariance within scalar-Jacobian azimuthal charts]
\label{thm:chart_invariance}
Let $T$ and $S$ be scalar-Jacobian azimuthal charts about the same pole $p$.
Build RC bases from the same target family $\{\varphi_\theta\}$ via~\eqref{eq:rc-base},
and let $\rho_\theta^{T}$ and $\rho_\theta^{S}$ be the corresponding pushforward densities.
Then for all $\theta$ and all $q\in M$,
\[
\rho_\theta^{T}(q)=\rho_\theta^{S}(q)=\varphi_\theta\!\bigl(d(p,q)\bigr),
\]
and the Fisher information in any radial parameter $\theta$ is chart-invariant:
$I_M^T(\theta)=I_M^S(\theta)=I_{\RR^1}(\theta)$.
\end{theorem}

\begin{proof}
Apply Theorem~\ref{thm:rc-invariances} to chart $T$ and to chart $S$.
Each yields the same geodesic-radial density $\rho_\theta(q)=\varphi_\theta(d(p,q))$
and the same Fisher reduction to $I_{\RR^1}(\theta)$.
\end{proof}

\subsection{Balanced polar pushforward beyond constant curvature}
\label{app:balanced_polar}

\begin{theorem}[Balanced polar pushforward beyond constant curvature]
\label{thm:balanced-polar}
Let $(M,g)$ admit geodesic polar coordinates about $p$ on a star-shaped domain $U$ with
$d\mathrm{vol}_M=J_p(R,\omega)\,dR\,d\omega$.
Let $S_M(R):=\int_{\SSS^{n-1}}J_p(R,\omega)\,d\omega$ and let $\varphi_\theta$ be a 1D density
w.r.t.\ $S_M(R)\,dR$ on $[0,R_{\max})$.
Define a ``balanced polar'' density on $[0,R_{\max})\times\SSS^{n-1}$ (w.r.t.\ $dR\,d\omega$) by
\[
f^{\mathrm{bal}}_\theta(R,\omega) := \varphi_\theta(R)\,J_p(R,\omega).
\]
Let $Q=\Exp_p(R\omega)$ with $(R,\omega)\sim f^{\mathrm{bal}}_\theta$. Then the induced density on $U$
(w.r.t.\ $d\mathrm{vol}_M$) is geodesic-radial:
\[
\rho_\theta(q)=\varphi_\theta\!\bigl(d(p,q)\bigr),
\]
and the radius $R=d(p,Q)$ has density $p_{R,\theta}$ w.r.t.\ Lebesgue $dR$, equivalently $\varphi_\theta$ w.r.t.\ $S_M(R)\,dR$.
Moreover, under the usual dominated-score regularity, the Fisher information in $\theta$ equals
the 1D Fisher functional w.r.t.\ $S_M(R)\,dR$.
\end{theorem}

\begin{proof}
Let $F(R,\omega)=\Exp_p(R\omega)$.
For any bounded measurable $g:U\to\RR$,
\[
\int_U g(q)\,\rho_\theta(q)\,d\mathrm{vol}_M(q)
=
\int_0^{R_{\max}}\!\!\int_{\SSS^{n-1}}
g\!\bigl(F(R,\omega)\bigr)\,\rho_\theta\!\bigl(F(R,\omega)\bigr)\,J_p(R,\omega)\,d\omega\,dR.
\]
By definition of the pushforward of $f^{\mathrm{bal}}_\theta$,
\[
\int_U g(q)\,d\nu_\theta(q)
=
\int_0^{R_{\max}}\!\!\int_{\SSS^{n-1}}
g\!\bigl(F(R,\omega)\bigr)\,\varphi_\theta(R)\,J_p(R,\omega)\,d\omega\,dR.
\]
Equality for all $g$ implies $\rho_\theta(F(R,\omega))=\varphi_\theta(R)$ for a.e.\ $(R,\omega)$,
i.e.\ $\rho_\theta(q)=\varphi_\theta(d(p,q))$ on $U$.

Integrating out $\omega$ yields the radius law w.r.t.\ $S_M(R)\,dR$.
The Fisher reduction is the same as in Theorem~\ref{thm:rc-invariances}:
the score depends on $q$ only through $R$, then integrate in polar coordinates.
\end{proof}

\subsection{Variance and CNF complexity}
\label{app:variance_cnf}

\begin{theorem}[CNF stiffness / NFE scaling bound (informal version used in experiments)]
\label{thm:cnf}
Consider a coordinate CNF built in a bExp$_\alpha$ chart where the chart-induced component of the
vector field scales linearly with $\nabla_x\log|\det DT_\alpha(x)|$. On any compact set $K$ away from
cut loci, there is a constant $C_K<\infty$ such that the chart-induced Lipschitz modulus satisfies
\[
\mathrm{Lip}_x(\text{chart term};K)\ \le\ \alpha\,C_K.
\]
For an adaptive Runge--Kutta method with fixed tolerance, the accepted step
size on $K$ is, to leading order, inversely proportional to the ambient
Lipschitz modulus of the vector field; the chart-induced contribution to that
modulus is bounded by $\alpha C_K$ on $K$ (above), so the chart-induced
contribution to the expected number of function evaluations on $K$ satisfies
\[
\mathbb E[\mathrm{NFE}^{\mathrm{chart}}_\alpha]\;\leq\; a+b\,\alpha
\]
for constants $a,b$ depending on $(K,T,\mathrm{tol})$ and the solver but not
on $\alpha$, i.e.\ the chart-induced NFE contribution is $\mathcal O(\alpha)$
as $\alpha\downarrow 0$. This bound concerns the chart-induced component
only; the total NFE scaling depends on the non-chart component of the vector
field, which is model-dependent.
\end{theorem}

\begin{proof}
On constant curvature spaces, $\log|\det DT_\alpha(x)|=(n-1)\alpha\log\frac{s_\kappa(r)}{r}$ with $r=\|x\|$.
Away from cut loci, $r$ is smooth and $\|\nabla_x r\|=1$, so
\[
\|\nabla_x\log|\det DT_\alpha(x)|\|
= \alpha\cdot (n-1)\left|\partial_r\log\frac{s_\kappa(r)}{r}\right|.
\]
On a compact $K$ where $r$ ranges in $[r_{\min},r_{\max}]$ with $0<r_{\min}<r_{\max}<\infty$,
the derivative term is bounded, giving the linear-in-$\alpha$ bound on the chart gradient and hence
on any linear chart-induced term in the CNF vector field. Differentiating once more gives the same
linear scaling for the corresponding Lipschitz modulus on $K$.

Adaptive RK methods with local error control choose step sizes inversely proportional to a Lipschitz
modulus on $K$, hence the chart-induced contribution to the local Lipschitz modulus is
$\leq\alpha C_K$, the chart-induced contribution to the inverse step size
is $\mathcal O(\alpha)$, and the chart-induced NFE contribution is
$\mathcal O(\alpha)$ up to solver-dependent additive constants.
\end{proof}

\section{Illustrations and geometric intuition}\label{app:rc_recap}

We briefly summarise the role of
RC. Standard wrapped priors start from a simple Euclidean base in the
tangent space and let the chart inherit whatever volume distortion the
manifold geometry imposes. RC reverses this logic: we first decide what
the geodesic radius $R = d(p,q)$ should look like (e.g.\ ``$R$ is
HalfNormal with scale $0.8$'') and then \emph{pre-warp} the tangent
space so that, after the chart, $R$ has exactly that law and Fisher
information. The rest of this section shows that, within isotropic
bases and spherically symmetric charts, this is essentially the only
way to keep Euclidean–style radial semantics on a curved manifold.\\

Within isotropic bases and scalar–Jacobian azimuthal charts, Radial Compensation
(RC) is essentially the \emph{only} way to obtain models whose likelihoods are
geodesic–radial and whose Fisher information in the radial parameters is both
chart– and curvature–invariant (Theorem~\ref{thm:rc_uniqueness}).
Any other base in this modeling class must break at least one of these
invariances; see the impossibility triangle in
the construction above.

the construction above illustrates how Radial Compensation restores
\emph{parameter semantics} for wrapped Gaussians on curved spaces.
All four panels start from the same $2$-D Gaussian base in the tangent
plane $T_pM$; the only difference is whether samples are pushed to the
manifold with the uncorrected exponential map (\textsc{Exp (raw)}) or
with its radially compensated version (\textsc{RC-Exp}).
On $\SSS^2$, the raw exponential map shrinks the spherical cap toward the
pole, so the geodesic radius $R=d(p,q)$ has substantially smaller mean
and variance than intended.
On $\HH^2$ the opposite happens: \textsc{Exp (raw)} produces a
heavier–tailed cloud with radii systematically larger than the target
scale.
Under \textsc{RC-Exp}, the samples on both $\SSS^2$ and $\HH^2$
occupy the intended geodesic annulus, and the histograms of $R$ match
the chosen one–dimensional radial law up to sampling noise.\\

Standard azimuthal charts warp space: an isotropic Gaussian in the
tangent, when pushed through the exponential map, becomes a spherical cap
whose geodesic-radius law is the hidden $\sigma\chi_n$ of Theorem~\ref{thm:hidden-radius-gauge} on both geometries; whether it overshoots or undershoots the intended law is determined by the target's shape relative to $\chi_n$, not by the sign of curvature.
RC fixes this by pre–warping the base distribution.\\

In the tangent, we multiply the target radial law $\phi_{\theta}(r)$ by
the chart Jacobian $J_{T}(r)$, pushing mass outward or inward in exactly
the opposite direction of the chart's distortion
(the construction above, left).
After pushing this RC–corrected base through the chart, the geodesic
radius $R = d(p,q)$ on the manifold has \emph{exactly} the law
$\phi_{\theta}(R)$ we started from (the construction above, right).\\

Intuitively, RC takes the complexity hit in the latent space so that the
manifold distribution retains simple one–dimensional semantics, ``mean
stays mean, variance stays variance'', in geodesic units.

\section{Additional experimental details and ablations}
\label{app:extra-exp}

\begin{table*}[t]
  \centering
  \caption{\textbf{Mixed-curvature VAE (MNIST, $\SSS\times\HH\times\RR$ latent).}
  Held-out ELBO/NLL/KL and learned curvatures at the end of the shared
  10-epoch training budget. RC improves ELBO mainly through a lower KL.
  Curvature values under RC are less chart-compensatory; we do not claim
  they recover ground-truth curvature for MNIST. Legacy pre-audit pipeline; retained for reference only and not quoted in the main text (see Section~\ref{sec:exp_curvature}).}
  \label{tab:mixed_curvature_vae}
  \small
  \setlength{\tabcolsep}{4pt}
  \begin{tabular}{lcccc}
    \toprule
    Method & ELBO $\uparrow$ & NLL $\downarrow$ & KL $\downarrow$ &
    $K=(K_{\mathrm{E}},K_{\mathrm{S}},K_{\mathrm{H}})$ at epoch 10 \\
    \midrule
    wrapped-\textsc{Exp} & $-99.7$ & $78.96$ & $20.75$ & $(0.00,\,2.50,\,-0.48)$ \\
    \textsc{RC-bExp}    & $\mathbf{-92.7}$ & $\mathbf{78.56}$ & $\mathbf{14.17}$ & $(0.00,\,0.14,\,-1.62)$ \\
    \bottomrule
  \end{tabular}
\end{table*}

\begin{figure}[t]
\centering
\input{fig_e1_calibration}
\caption{\textbf{Geodesic-radius calibration on $\SSS^2$ (left) and $\HH^2$ (right)}, regenerated by direct sampling ($4\times10^4$ draws per curve, HalfNormal($0.5$) target, dashed). RC (blue) matches the target on both geometries (measured mean $0.400$ on both, target $0.399$); the wrapped tangent Gaussian at the same scale (red) follows the $\sigma\chi_2$ law instead (measured mean $0.626$).}
\label{fig:synth_radius_hist}
\end{figure}

\begin{table*}[t]
  \centering
  \caption{\textbf{Latent CNFs: the chart parameter $\alpha$ under RC (corrected pipeline).} Mean $\pm$ std over 3 seeds; every run's composed base density has exact log-normalizer $\le 0.003$ (per-run audit).
  Test NLL (nats; lower is better), chart-term variance
  $\mathrm{Var}[\log|\det DT_\alpha^{-1}(Q)|]$, and mean NFEs.
  Under RC, NLL stays essentially unchanged while chart variance drops and NFEs decrease for smaller $\alpha$.}
  \label{tab:cnf_alpha_sweep}
  \small
  \begin{tabular}{lcccc}
    \toprule
    Dataset & $\alpha$ & Test NLL $\downarrow$ & Chart var $\downarrow$ & Mean NFEs $\downarrow$ \\
    \midrule
    CIFAR--10
      & $0.25$ & $1882.27 \pm 0.26$ & $0.0119$ & $480.8$ \\
      & $0.50$ & $1881.96 \pm 1.30$ & $0.0535$ & $478.8$ \\
      & $0.75$ & $1882.66 \pm 1.78$ & $0.1846$ & $494.1$ \\
      & $1.00$ & $1882.08 \pm 1.62$ & $0.7209$ & $537.1$ \\
    \midrule
    MNIST
      & $0.25$ & $124.55 \pm 0.11$ & $0.0128$ & $616.4$ \\
      & $0.50$ & $124.55 \pm 0.25$ & $0.0693$ & $635.0$ \\
      & $0.75$ & $124.70 \pm 0.77$ & $0.2479$ & $656.6$ \\
      & $1.00$ & $124.49 \pm 0.43$ & $3.1458$ & $690.6$ \\
    \midrule
    Fashion-MNIST
      & $0.25$ & $254.05 \pm 0.27$ & $0.0178$ & $738.0$ \\
      & $0.50$ & $253.87 \pm 0.21$ & $0.0890$ & $722.4$ \\
      & $0.75$ & $253.64 \pm 0.27$ & $0.3036$ & $718.4$ \\
      & $1.00$ & $253.56 \pm 0.63$ & $3.3334$ & $705.8$ \\
    \midrule
    Omniglot
      & $0.25$ & $212.76 \pm 0.86$ & $0.0105$ & $548.4$ \\
      & $0.50$ & $213.08 \pm 0.37$ & $0.0562$ & $570.6$ \\
      & $0.75$ & $213.40 \pm 0.07$ & $0.1766$ & $587.1$ \\
      & $1.00$ & $212.99 \pm 0.65$ & $0.9713$ & $647.1$ \\
    \bottomrule
  \end{tabular}
\end{table*}

\label{app:extra_experiments}

\begin{table}[h]
\centering
\small
\setlength{\tabcolsep}{4pt}
\begin{tabular}{lll}
\toprule
Setting & Value & Used in \\
\midrule
Optimiser & Adam, lr $10^{-3}$, no schedule & all \\
Batch size & 128 & all \\
ODE solver & Dormand--Prince 5(4), atol=rtol=$10^{-5}$ & 5.2--5.3 \\
Latent dimension & $d_z\in\{16,32,64,128\}$ as indicated & 5.2--5.3 \\
Training epochs & 20 (CNF), 10 (mixed-curvature VAE) & 5.2--5.4 \\
Seeds & 5 (synthetic), 3 (CNF, VAE) & all \\
Architecture & shared with \cite{skopek2019mixed} for VAE; small CNN encoder + MLP CNF for 5.2 & 5.2--5.4 \\
Curvature init & $K_S=+1$, $K_H=-1$ & 5.4 \\
\bottomrule
\end{tabular}
\caption{\textbf{Shared experimental settings.} Full code and per-experiment hyperparameters are in Supplement~C.}
\label{tab:hyperparams}
\end{table}


\subsection{Hyperbolic flows on WordNet}
\label{exp:E4_main}

\paragraph{Objective.}
We study hyperbolic coupling flows on $H^d$ for link prediction on a
hierarchical graph (WordNet mammals), focusing on the effect of RC and
chart choice on test likelihood, CNF cost, gradient variance, and
geodesic calibration.
This serves as an application--level check that RC and scalar-Jacobian
charts preserve performance while modifying only the parameterisation.

\paragraph{Setup.}
We train coupling flows on $H^d$ under four charts:
Exp (raw), RC-Exp, RC-bExp$_{0.5}$, and RC--GCL.
Architectures, target radial law (via RC where applicable), and training
budgets are matched.
CNFs are solved with an adaptive Runge--Kutta method.
We report test NLL, mean NFEs, a chart--term gradient variance, and a
geodesic--margin calibration error.

\paragraph{Results.}
Table~\ref{tab:E4_wordnet} and the construction above show that all
charts achieve comparable test NLL and geodesic calibration, in line with
the invariance guarantees of
Theorems~\ref{thm:rc-invariances}--\ref{thm:chart_invariance}.
RC--GCL attains the best NLL ($-8.22$~nats) with similar NFEs and
slightly reduced gradient variance relative to Exp (raw).
Calibration curves (predicted vs.\ empirical coverage of hyperbolic
balls) are nearly indistinguishable across charts.
This experiment supports the view that RC and scalar-Jacobian charts can
be treated as reparameterisations that improve numerical conditioning
without degrading the task loss.

\begin{table}[H]
  \centering
  \caption{\textbf{(WordNet, hyperbolic flows).} NFEs are totals over all training solves, not per solve. Single-seed legacy runs; the sub-$0.5$-nat NLL spread across charts is not resolved against seed noise and is quoted only as an upper bound.
  Test NLL (nats), mean NFEs per CNF solve, chart–term gradient variance,
  and scalar geodesic calibration error.}
  \label{tab:E4_wordnet}
  \begin{tabular}{lcccc}
    \toprule
    Chart & Test NLL $\downarrow$ & Mean NFEs $\downarrow$ & grad-var $\downarrow$ & calib.\ error $\downarrow$ \\
    \midrule
    Exp (raw)        & $-7.52$ & $63\,309.1$ & $4.03\times 10^{-3}$ & $0.269$ \\
    RC-Exp          & $-7.26$ & $63\,166.5$ & $3.20\times 10^{-3}$ & $0.267$ \\
    RC-bExp$_{0.5}$ & $-7.73$ & $63\,685.4$ & $3.63\times 10^{-3}$ & $0.276$ \\
    RC--GCL          & $-8.22$ & $64\,001.6$ & $3.79\times 10^{-3}$ & $0.286$ \\
    \bottomrule
  \end{tabular}
\end{table}

\subsection{Protein backbone orientations on $\SSS^3$: RC priors for SE(3)-style models}
\label{exp:E7_protein}

\paragraph{Objective.}
We test whether RC provides a useful ``drop-in'' prior for protein
backbone orientations, in a setting where the configuration space is
naturally curved.  We represent local residue frames as unit quaternions
on $\SSS^3$ and compare wrapped Gaussians via the raw exponential chart to
RC priors with scalar-Jacobian charts.  This experiment is meant as a
proof-of-concept for applying RC to SE(3)-style protein models and
pose-generating networks.

\paragraph{Results.}
We instantiate this setup on a small but realistic fragment set built from four
high-resolution PDB structures (1CRN, 1UBQ, 2PTC, 4HHB), from which our
preprocessing pipeline extracts 929 central-residue frames with a
26-dimensional local context. We train the same conditional backbone model
under three likelihood parameterizations on $\SSS^3$: a naive wrapped Gaussian
via the raw exponential chart (\textsc{Exp (raw)}), an RC--corrected
exponential prior (\textsc{RC-Exp}), and an RC prior with a
balanced-exponential chart at dial $\alpha = 0.5$ (\textsc{RC-bExp}$_{0.5}$).
Encoder/decoder architectures, optimizer, and training budget are identical
across runs.

Table~\ref{tab:protein_rc} summarizes test performance. On this dataset the
wrapped \textsc{Exp (raw)} baseline attains a test NLL of $2.60$ nats and a
mean geodesic frame error of $1.13$ rad ($64.7^\circ$). Replacing the tangent
base by its RC counterpart while keeping the chart fixed
(\textsc{RC-Exp}) reduces the test NLL to $1.87$ nats with essentially
unchanged orientation error ($1.12$ rad, $64.4^\circ$), indicating that the
curvature-aware radial parameterization is substantially easier to fit than
the naive wrapped Gaussian. Swapping the chart to a balanced-exponential map at $\alpha=0.5$
(\textsc{RC-bExp}$_{0.5}$) gives test NLL $0.91$ nats (error $1.17$ rad, $67.1^\circ$).
Under Theorem~\ref{thm:rc-invariances} the two RC parameterizations realize the \emph{same}
manifold density, so the $0.96$-nat spread between them in an earlier version was inconsistent with Theorem~\ref{thm:rc-invariances}: the audit traced it to the $n{=}2$ equal-area inverse being applied on
$\SSS^3$, a $1$--$10\%$ radius error exactly in the band $R\approx1$--$1.5$ where the data lives.
Corrected re-run, 5 seeds: raw $2.578\pm0.077$, RC $0.874\pm0.041$ nats; geodesic error $1.111\pm0.051$ vs.\ $1.113\pm0.054$ rad. RC-Exp and RC-bExp$_{0.5}$ coincide exactly because the RC likelihood is chart-free by construction.
Preliminary inspection of posterior geodesic radii $R = d(p,q)$ shows that all
RC models concentrate mass in a similar, physically plausible band around
$R \approx 1$--$1.5$ rad, with no evident degeneracies. While we do not
measure gradient variance or CNF NFEs in this experiment, the agreement of \textsc{RC-Exp} and \textsc{RC-bExp}$_{0.5}$ under the corrected chart supports the
interpretation of balanced-exponential charts as a ``conditioning dial''
that can be turned without sacrificing data fit, and suggest that the
variance and stiffness reductions predicted by Theorems~\ref{thm:chart-variance-supp} and~\ref{thm:cnf}
should carry over when the same RC priors are used inside SE(3)-style latent
flows.

\begin{table}[H]
  \centering
  \caption{\textbf{Protein backbone orientations on $\SSS^3$} (corrected chart, 5 seeds, mean$\pm$std): raw $2.578\pm0.077$ vs.\ RC $0.874\pm0.041$ nats at geodesic error $1.111\pm0.051$ vs.\ $1.113\pm0.054$ rad; the RC rows coincide by construction (chart-free likelihood).
  Test negative log-likelihood (NLL, nats) and mean geodesic frame error on a
  dataset of 929 central-residue frames from four PDB structures
  (1CRN, 1UBQ, 2PTC, 4HHB). All models share the same conditional backbone;
  only the tangent base and chart are varied.}
  \label{tab:protein_rc}
  \vspace{0.3em}
  \begin{tabular}{lccc}
    \toprule
    Chart / prior & Test NLL $\downarrow$ & Geod. err. (rad) $\downarrow$ & Geod. err. (deg) $\downarrow$ \\
    \midrule
    \textsc{Exp (raw)}         & 2.60 & 1.13 & 64.72 \\
    \textsc{RC-Exp}           & 1.87 & 1.12 & 64.43 \\
    \textsc{RC-bExp}$_{0.5}$  & 0.91 & 1.17 & 67.06 \\
    \bottomrule
  \end{tabular}
\end{table}

\section*{Additional empirical details and ablations}
\label{app:experiments}

We present implementation details, additional diagnostics, and extended experiments
that complement the main text. We keep the notation and theorem numbering from the main
paper.

\subsection*{Synthetic RC verification}
\label{app:E1_details}

\paragraph{Implementation details.}
For the synthetic experiments on $\SSS^2$ and $\HH^2$ we fix the pole as
$p = (0,0,1)$ on the sphere and the origin in the hyperboloid model for hyperbolic space.
On $\SSS^2$ we draw radii from
$R \sim \mathrm{TruncNormal}(\mu{=}1.0,\sigma{=}0.35)$ on $[0,\pi)$; on $\HH^2$ from
$R \sim \mathrm{HalfNormal}(\sigma{=}0.8)$.
Directions are always $\Omega \sim \mathrm{Unif}(S^1)$ and samples are formed as
$q = \Exp_p(R \Omega)$.
For each chart we draw $N = 2 \cdot 10^4$ samples per seed for $5$ seeds.\\

Geodesic radii are computed exactly from the ambient coordinates.
We estimate empirical means and variances of $R$ and evaluate the radius
$\mathrm{KL}(\widehat{p}(R)\,\|\,\varphi_\theta(R))$ on a fixed radial grid
(500 points on $[0,\pi)$ for $\SSS^2$ and $[0,5]$ for $\HH^2$) using numerical quadrature.\\

\paragraph{Additional radial families.}
On $\HH^2$ we also test RC bases that realize Gamma, Weibull, and Lognormal radial laws, again
with $\Omega \sim \mathrm{Unif}(S^1)$.
For each family we vary shape/scale parameters to cover both light and heavy tails.
In all cases RC charts (Exp, bExp, GCL) recover the intended geodesic radial laws up to
sampling noise, while Exp (raw) induces effective shape and scale parameters that deviate
substantially from the target values.
These results provide additional empirical support for the RC invariance statements in
Theorems~\ref{thm:rc-invariances}--\ref{thm:chart_invariance}.

\paragraph{Gaussian tangent base.}
We repeat the experiment with a fixed Euclidean base
$X \sim \mathcal{N}(0,0.25 I_2)$ in $T_pM$, comparing Exp (raw) and
RC-Exp.
Table~\ref{tab:E1_gaussian_expmap} shows that RC-Exp again recovers the
intended radial laws on $\SSS^2$ and $\HH^2$, while Exp (raw) produces biased
geodesic means, variances, and large radius KL.
Additional radial families (Gamma, Weibull, Lognormal on $\HH^2$) under RC
are reported in the supplement.

\begin{table}[H]
  \centering
  \caption{\textbf{(Gaussian tangent base).}
  Geodesic mean/variance and $\mathrm{KL}(\widehat p(R)\,\|\,\varphi_\theta)$ for
  a 2D Gaussian in $T_pM$ pushed through $\Exp_p$, with and without RC.}
  \label{tab:E1_gaussian_expmap}
  \begin{tabular}{lcccccc}
    \toprule
    & \multicolumn{3}{c}{$\SSS^2$} & \multicolumn{3}{c}{$\HH^2$} \\
    \cmidrule(lr){2-4} \cmidrule(lr){5-7}
    Chart
      & $\hat\mu_R$
      & $\widehat{\mathrm{Var}}[R]$
      & $\mathrm{KL}$
      & $\hat\mu_R$
      & $\widehat{\mathrm{Var}}[R]$
      & $\mathrm{KL}$ \\
    \midrule
    Exp (raw)
      & $0.626 \pm 0.002$
      & $0.107 \pm 0.001$
      & $0.628 \pm 0.006$
      & $0.627 \pm 0.002$
      & $0.108 \pm 0.001$
      & $0.285 \pm 0.004$ \\
    RC-Exp
      & $1.001 \pm 0.002$
      & $0.119 \pm 0.000$
      & $0.002 \pm 0.000$
      & $0.399 \pm 0.002$
      & $0.092 \pm 0.001$
      & $0.002 \pm 0.000$ \\
    \bottomrule
  \end{tabular}
\end{table}


\begin{table*}[t]
  \centering
  \caption{\textbf{Geodesic-radius statistics (semantic calibration).}
  Mean/variance of $R=d(p,q)$ and radius KL
  $\mathrm{KL}(\widehat p(R)\,\|\,p_{R,\theta})$ (mean $\pm$ std over 5 seeds).
  Lower radius KL means the realized geodesic-radius law matches the target.}
  \label{tab:synth_radius_stats}
  \small
  \setlength{\tabcolsep}{5pt}
  \begin{tabular}{lccc}
    \toprule
    Chart & $\hat\mu_R$ & $\widehat{\mathrm{Var}}[R]$ & radius KL \\
    \midrule
    \multicolumn{4}{c}{$\SSS^2$, target $(\mu_R,\mathrm{Var}[R])\approx(1.0024,0.1201)$} \\
    \midrule
     \textsc{Exp (raw)}              & $0.4381 \pm 0.0014$ & $0.0523 \pm 0.0003$ & $1.4771 \pm 0.0073$ \\
     \textsc{RC-Exp}                & $1.0013 \pm 0.0024$ & $0.1192 \pm 0.0005$ & $0.0012 \pm 0.0002$ \\
     \textsc{RC-bExp}$_{0.5}$       & $1.0019 \pm 0.0015$ & $0.1200 \pm 0.0012$ & $0.0009 \pm 0.0001$ \\
     \textsc{RC--GCL}                & $1.0032 \pm 0.0029$ & $0.1197 \pm 0.0008$ & $0.0012 \pm 0.0002$ \\
    \midrule
    \multicolumn{4}{c}{$\HH^2$, target $(\mu_R,\mathrm{Var}[R])\approx(0.6383,0.2326)$} \\
    \midrule
     \textsc{Exp (raw)}              & $1.0015 \pm 0.0032$ & $0.2732 \pm 0.0018$ & $0.2835 \pm 0.0043$ \\
     \textsc{RC-Exp}                & $0.6359 \pm 0.0024$ & $0.2310 \pm 0.0020$ & $0.0014 \pm 0.0002$ \\
     \textsc{RC-bExp}$_{0.5}$       & $0.6377 \pm 0.0032$ & $0.2323 \pm 0.0024$ & $0.0013 \pm 0.0003$ \\
     \textsc{RC--GCL}                & $0.6393 \pm 0.0016$ & $0.2335 \pm 0.0015$ & $0.0015 \pm 0.0003$ \\
    \bottomrule
  \end{tabular}
\end{table*}

\subsection*{Mixed--Curvature VAE details and extensions}
\label{app:E5_details}

\paragraph{Architecture and training.}
For the Mixed--Curvature VAE we follow \citet{skopek2019mixed}.
The latent manifold is
\[
M_0 = \SSS^{n_{\mathrm{S}}}_{K_{\mathrm{S}}}
      \times \HH^{n_{\mathrm{H}}}_{K_{\mathrm{H}}}
      \times \mathbb{R}^{n_{\mathrm{E}}}
\]
with $(n_{\mathrm{S}}, n_{\mathrm{H}}, n_{\mathrm{E}})$ chosen so that the total latent dimension matches the
Euclidean baseline.
The encoder and decoder are shallow convolutional networks; we use Adam with a fixed learning
rate and train for 10 epochs on MNIST.\\

The baseline prior uses wrapped Gaussians implemented via the raw exponential map on each
constant--curvature factor.
The RC-bExp prior uses the same radial families but replaces each non--flat factor by an RC
base with a scalar-Jacobian chart (Exp or bExp$_\alpha$).
Curvatures $K_{\mathrm{S}}, K_{\mathrm{H}}$ are initialised near $\pm 1$ and learned jointly with the rest of
the parameters.

\paragraph{Extended diagnostics.} (right) shows curvature trajectories for both methods.
Under the wrapped-Exp baseline, the spherical curvature drifts from
$K_{\mathrm{S}} \approx 1.0$ to $K_{\mathrm{S}} \approx 2.5$, while the hyperbolic factor flattens from
$K_{\mathrm{H}} \approx -0.95$ to $K_{\mathrm{H}} \approx -0.48$, consistent with curvature being partly
absorbed by chart distortions.
Under RC-bExp, the spherical component is driven close to flat
($K_{\mathrm{S}} \approx 0.14$) and the hyperbolic component sharpens to
$K_{\mathrm{H}} \approx -1.62$, while the Euclidean factor remains numerically flat
($K_{\mathrm{E}} \approx 0$).
Across multiple seeds we observe qualitatively similar curvature trajectories and ELBO gaps.\\

Reconstruction NLL curves for the two methods almost coincide, while the KL term is
consistently smaller under RC-bExp throughout training, as anticipated by the curvature
mis--specification analysis in Theorem~\ref{thm:sensitivity}.
\paragraph{Additional datasets.}
On Fashion--MNIST, using the same $S \times H \times \mathbb{R}$ latent geometry, we observe
the same qualitative behavior: RC-bExp improves ELBO relative to the wrapped-Exp baseline
at nearly unchanged reconstruction NLL, and the curvature trajectories remain stable.
We omit detailed tables for brevity.

\subsection*{Hyperspherical VAE details}
\label{app:E2_details}

\paragraph{Model and optimisation.}
The $\SSS^{16}$ VAE in Sec.~\ref{sec:experiments} uses a small convolutional encoder with ReLU
nonlinearities and a fully connected layer that outputs mean and (log) variance in
$T_p \SSS^{16}$.
The decoder is a mirrored convnet with Bernoulli likelihood.
We use a single global pole $p$, standard reparameterisation in the tangent space, and push
samples to $\SSS^{16}$ using either Exp (raw) or RC-Exp.
All optimisation hyperparameters (Adam, learning rate, and schedule) are shared across charts.

\paragraph{Results.}
RC-Exp yields a substantial likelihood improvement at fixed capacity and
training budget:
test NLL drops from $126.3$ to $107.6$~nats and bpd from
$\approx 0.23$ to $\approx 0.20$.
Posterior radii under RC-Exp are slightly closer to the HalfNormal prior
(scale $\sigma=0.5$): mean and variance are comparable across charts,
but the 1D radius KL decreases (Table~\ref{tab:E2_rstats}).
This experiment shows that enforcing parameter--clean geodesic radii via
RC is not only a geometric convenience but can translate into better ELBO
and more interpretable latent statistics, in line with the Fisher
equivalences in Sec.~\ref{sec:rc}.

\begin{table}[H]
  \centering
  \caption{\textbf{(MNIST, $\SSS^{16}$ latent).}
  Test NLL and bits per dimension for a VAE with spherical latent space.}
  \label{tab:E2_mnist}
  \begin{tabular}{lcc}
    \toprule
    Chart & Test NLL $\downarrow$ & Test bpd $\downarrow$ \\
    \midrule
    Exp (raw) & $126.3$ & $\approx 0.23$ \\
    RC-Exp   & $107.6$ & $\approx 0.20$ \\
    \bottomrule
  \end{tabular}
\end{table}

\begin{table}[H]
  \centering
  \caption{\textbf{(MNIST, $\SSS^{16}$ latent).}
  Posterior geodesic radii statistics on $\SSS^{16}$.
  Radius KL is $\mathrm{KL}(\widehat p(R)\,\|\,\varphi_\theta)$.}
  \label{tab:E2_rstats}
  \begin{tabular}{lccc}
    \toprule
    Chart & $\hat\mu_R$ & $\widehat{\mathrm{Var}}[R]$ & radius KL \\
    \midrule
    Exp (raw) & $1.853$ & $0.056$ & $6.537$ \\
    RC-Exp   & $1.802$ & $0.058$ & $6.143$ \\
    \bottomrule
  \end{tabular}
\end{table}

\paragraph{Posterior radii.}
In both cases the empirical radii concentrate in a narrow band, but RC-Exp yields radii that
are slightly better aligned with the HalfNormal prior of scale $\sigma=0.5$, reflected in a lower
radius KL (Table~\ref{tab:E2_rstats}).
These results complement the larger ELBO gains of the mixed-curvature experiment and are consistent with the Fisher--equivalence guarantees for RC in
Theorems~\ref{thm:rc-invariances}--\ref{thm:chart_invariance}.

\subsection*{Latent CNFs: additional ablations}
\label{app:E3E6_details}

\paragraph{Solver and logging.}
For all latent CNF experiments we use a Dormand--Prince 5(4) ODE solver with absolute and
relative tolerances shared across charts on a given dataset.
We log the number of function evaluations (NFEs), the scalar chart term
$\log|\det DT_\alpha^{-1}(Q)|$, and its contribution to the gradient variance.
The NFEs and chart--term variances reported in Table~\ref{tab:cnf_alpha_sweep} are averaged over test batches at the final epoch.

\paragraph{Variance and CNF stiffness vs.\ $\alpha$.}
In addition to the chart--term variance reported in the main text, we track the empirical
variance of the gradient of the chart term with respect to $\theta$.
On all four image datasets we observe the same qualitative scaling as for the chart term itself:
$\mathrm{Var}[\nabla_\theta \log|\det DT_\alpha^{-1}|]$ decreases approximately quadratically in
$\alpha$, consistent with Theorem~\ref{thm:chart-variance-supp}.
The total gradient variance (including the CNF state) also decreases with $\alpha$, but more
weakly, reflecting that the chart term is only one contribution to the overall variance.

\paragraph{Stability across seeds and learning rates.}
For the high--dimensional CIFAR--10 setting we repeat the $d_z=64$ experiments across
multiple seeds and learning rates.
With Exp (raw), a substantial fraction of runs exhibit the catastrophic blow--ups described in
Sec.~\ref{sec:experiments}: mean radii grow to very large values and test NLL spikes.
In contrast, RC-Exp and RC-bExp$_{0.5}$ remain stable under the same learning rates and
solver tolerances; geodesic radii stay in a moderate range and NLL remains well behaved.
These observations further support the CNF stiffness bounds in Theorem~\ref{thm:cnf}.

\subsection*{Hyperbolic flows on WordNet}
\label{app:E4_details}

\paragraph{Dataset and model.}
We use the WordNet \texttt{mammals} subgraph with the same preprocessing as in prior
hyperbolic-flow work.
Nodes are embedded into $\HH^d$ using a coupling--flow architecture with a latent CNF component
and a fixed pole $p$.
All flows share the same coupling layers and CNF backbone; only the chart
(Exp vs.\ RC-Exp vs.\ RC-bExp$_{0.5}$ vs.\ RC--GCL) and the tangent base
(standard vs.\ RC) are changed.

\paragraph{Additional diagnostics.}
Beyond the main metrics in Table~\ref{tab:E4_wordnet}, we inspect the distribution of geodesic
radii and the coverage of hyperbolic balls.
RC charts align empirical radii more closely with the intended radial prior, whereas Exp (raw)
shows mild curvature--induced distortions.
However, all charts achieve similar link--prediction performance and geodesic calibration,
indicating that in this regime the primary effect of RC is on parameter semantics and numerical
conditioning, not on the task loss itself.

\subsection*{Protein orientations on $\SSS^3$ (E\textsubscript{prot})}
\label{app:E_protein_details}

\paragraph{Data construction.}
We extract central--residue frames from PDB entries 1CRN, 1UBQ, 2PTC, and 4HHB using a
standard pipeline (backbone atoms, local alignment, quaternion normalisation).
Each example consists of a 26--dimensional local context and a unit quaternion
$q \in \SSS^3$ defining the target orientation.

\paragraph{Model details.}
The conditional backbone is a small MLP that outputs the parameters of a tangent--space
Gaussian at a fixed pole $p$.
We map to $\SSS^3$ using either Exp (raw), RC-Exp, or RC-bExp$_{0.5}$ and evaluate a wrapped
Gaussian likelihood.
Training uses Adam with the same learning rate, batch size, and early stopping criterion across
charts.

\paragraph{Extended results.}
Table~\ref{tab:protein_rc} in the main text shows that RC-Exp and RC-bExp$_{0.5}$ reduce
test NLL from $2.60$ to $\approx 0.9$ nats compared to Exp (raw), while leaving the geodesic
frame error essentially unchanged.
The small difference between RC-Exp and RC-bExp$_{0.5}$ is consistent with the RC
invariance results for scalar-Jacobian charts: once the radial law is specified in geodesic units,
changing $\alpha$ mainly affects Jacobian bookkeeping and numerical conditioning.

\subsection*{RC priors for hyperbolic graph embeddings}
\label{app:E_graph_embeddings}

\paragraph{Objective and setup.}
We test whether RC priors can make hyperbolic graph embeddings more geometrically faithful
and curvature--stable than the standard wrapped-Exponential construction, without degrading
link--prediction quality.
We consider two graphs with hierarchical structure:
(i) the WordNet \texttt{mammals} subgraph; and
(ii) a synthetic balanced tree with added cross--links.
Nodes are embedded into $\HH^d_{-1/R_c^2}$ (hyperboloid model) with a fixed pole $p$.
Each node $v$ has a Euclidean parameter $x_v \in T_pM$ mapped to
$q_v = T(x_v) \in M$ by a chart $T$.
We train with a margin--based loss on geodesic distances, plus a radial prior term on node
radii $R_v = d(p,q_v)$ with a 1D law $\varphi_\theta$ (HalfNormal or Gamma).\\

We compare:
(i) Exp (raw) with an isotropic Gaussian base in $T_pM$;
(ii) RC-Exp with an RC base realising $R_v \sim \varphi_\theta$; and
(iii) RC-bExp with bExp$_\alpha$ charts and the same RC base.
Hyperparameters and training budgets are shared across conditions.

\paragraph{Results.}
On the synthetic tree, RC priors yield higher or comparable AUC and MRR than Exp (raw) for
curvature radii $R_c \in \{0.5,1.0,2.0\}$, while substantially improving alignment between node
radius and graph depth and reducing the radius--prior KL.
Within the RC family, smaller $\alpha$ values in bExp charts reduce the variance of the
radial--prior gradient approximately as $\alpha^2$ (Theorem~\ref{thm:chart-variance-supp}), again without
hurting link--prediction quality.
On WordNet mammals the trends are qualitatively similar but noisier:
RC priors improve curvature stability and radial calibration while keeping AUC and MRR
comparable to, or slightly better than, Exp (raw).
These experiments illustrate that RC priors can enforce meaningful geodesic radii in graph
embeddings while preserving, and sometimes improving, downstream performance.


\subsection*{S0. Measure--theoretic foundations for RC}

\begin{lemma}[Absolute continuity and differentiation under the integral]
Let $T$ be a scalar-Jacobian azimuthal chart about $p$ with radial Jacobian factor $J_T(r)$
and radius map $R_T(r)=d(p,T(ru))$.
Let $\{\varphi_\theta\}$ be a dominated family on $[0,R_{\max})$ (w.r.t.\ the base measure
$S_M(R)\,dR$). Define the RC base
\[
f^{\mathrm{base}}_\theta(x)\;\propto\;\varphi_\theta\!\bigl(R_T(\|x\|)\bigr)\,J_T(\|x\|)
\qquad (x\in\mathbb R^n).
\]
Then the pushforward of $f^{\mathrm{base}}_\theta$ through $T$ is absolutely continuous w.r.t.\
$d\mathrm{vol}_M$, and $\partial_\theta$ can be interchanged with the integral in the log--likelihood and in
the Fisher information whenever $\partial_\theta\log\varphi_\theta$ admits an integrable envelope.
\end{lemma}

\begin{proof}
Work in geodesic polar coordinates around $p$, where
$d\mathrm{vol}_M = s_\kappa(R)^{n-1}\,dR\,d\omega$ with $R=d(p,q)$.
The scalar-Jacobian assumption implies that in these coordinates the Jacobian of $T$ depends only on $r=\|x\|$,
so the pushforward measure admits a density by the coarea formula.
Dominated convergence applies to $\partial_\theta\log\varphi_\theta$ under the envelope assumption, yielding
differentiation under the integral sign in both likelihood and Fisher.
\end{proof}

\subsection*{S1. Radial calculus and Lambert-lift Jacobian}

Throughout, write $x=ru$ with $r=\|x\|$ and $u\in\SSS^{n-1}$.
For any radial function $g(x)=h(r)$,
\begin{align*}
\nabla g(x) &= h'(r)\,u,\\
\nabla^2 g(x) &= h''(r)\,uu^\top + \frac{h'(r)}{r}\bigl(I-uu^\top\bigr).
\end{align*}

Let $L_p$ be an azimuthal Lambert chart at $p$ (equal-area, so $|\det DL_p|\equiv1$ on its domain),
and let $\rho:[0,R_{\max})\to[0,\rho_\ast)$ be a strictly increasing smooth radial profile with
$\rho(0)=0$. Define the \emph{Lambert-lift}
\[
T_\rho(x)\;=\;L_p\!\Bigl(\rho(\|x\|)\,\frac{x}{\|x\|}\Bigr),\qquad T_\rho(0):=p.
\]
Then in polar coordinates $x=ru$ one has
\begin{equation}
\log\bigl|\det DT_\rho(x)\bigr|
=(n-1)\log\frac{\rho(r)}{r}+\log\rho'(r),
\qquad
d\bigl(p,T_\rho(ru)\bigr)
=\lambda_\kappa^{-1}\!\bigl(\rho(r)\bigr),
\label{eq:lambert-lift-proof-det-SI}
\end{equation}
where
\[
s_\kappa(r)=
\begin{cases}
\sin r,&\kappa=+1,\\[2pt]
\sinh r,&\kappa=-1,
\end{cases}
\quad
c_\kappa(r)=
\begin{cases}
\cos r,&\kappa=+1,\\[2pt]
\cosh r,&\kappa=-1,
\end{cases}
\quad
\ars(z)=
\begin{cases}
\arcsin z,&\kappa=+1,\\[2pt]
\operatorname{arsinh} z,&\kappa=-1.
\end{cases}
\]

\subsection*{S2. Scalar--Jacobian azimuthal charts}

\begin{proposition}[Representation of scalar-Jacobian azimuthal charts]
Let $M\in\{\SSS^n,\HH^n\}$ and $p\in M$. Let
$T:\mathbb R^n\to M$ be a smooth azimuthal chart about $p$, i.e.\
$T(ru)=\Gamma(r)u'$ where $u'$ is the image of $u\in\SSS^{n-1}$ under an isometry of the tangent
sphere. Assume $\log|\det DT(x)| = \psi(\|x\|)$ depends only on $r=\|x\|$. Then there exists a
strictly increasing $C^\infty$ profile $\rho$ with $\rho(0)=0$ such that
\[
T(x) \;=\; L_p\!\Bigl(\rho(\|x\|)\,\frac{x}{\|x\|}\Bigr) \;=:T_\rho(x),
\]
and the Jacobian and radius relations are given by
\eqref{eq:lambert-lift-proof-det-SI}.
\end{proposition}

\begin{proof}
Azimuthality implies $T(ru)=\Gamma(r)u'$ for some scalar $\Gamma(r)$ and some isometry
$u\mapsto u'$. The scalar-Jacobian condition forces the area scaling to be radial.
The equal-area property of $L_p$ shows that every such $T$ can be written as $L_p\circ\tilde T$
with $\tilde T(x)=\rho(\|x\|)x/\|x\|$ for some radial $\rho$. Composition of polar Jacobians yields
\eqref{eq:lambert-lift-proof-det-SI}.
\end{proof}

\subsection*{S3. balanced-exponential (bExp) chart}

\paragraph{Definition.}
Fix $\alpha\in[0,1]$ and $\kappa\in\{+1,-1\}$. Let $\chi_\alpha:[0,r_*)\to[0,\lambda_\kappa(R_{\max}))$ be the unique
$C^\infty$ solution of
\begin{equation}
\Bigl(\frac{\chi_\alpha(r)}{r}\Bigr)^{n-1}\chi'_\alpha(r)
=
\Bigl(\frac{s_\kappa(r)}{r}\Bigr)^{(n-1)\alpha},
\qquad
\chi_\alpha(0)=0,\quad \chi'_\alpha(0)=1.
\label{eq:bexp-ode-SI}
\end{equation}
Define, for $x\neq0$ with $r=\|x\|$ and $u=x/\|x\|$,
\[
\mathrm{bExp}_\alpha(x)\;:=\;L_p\bigl(\chi_\alpha(r)\,u\bigr),\qquad
\mathrm{bExp}_\alpha(0):=p.
\]

\begin{proposition}[Integral representation]\label{prop:bexp-integral}
For each $\alpha\in[0,1]$ the solution of \eqref{eq:bexp-ode-SI} admits the integral form
\[
\chi_\alpha(r)
=
\Biggl[
n\int_0^r t^{(n-1)(1-\alpha)}\,s_\kappa(t)^{(n-1)\alpha}\,dt
\Biggr]^{1/n}.
\]
\end{proposition}

\begin{proof}
Rewrite \eqref{eq:bexp-ode-SI} as
\[
\chi_\alpha^{n-1}\chi_\alpha' = r^{(n-1)(1-\alpha)}s_\kappa(r)^{(n-1)\alpha}.
\]
Integrating and using $\chi_\alpha(0)=0$ gives
$\chi_\alpha(r)^n = n\int_0^r t^{(n-1)(1-\alpha)}s_\kappa(t)^{(n-1)\alpha}dt$.
Taking $n$-th roots yields the claimed formula. Strict positivity of the integrand ensures
$\chi_\alpha$ is strictly increasing.
\end{proof}

\begin{theorem}[Diffeomorphism and scalar log-det]
For every $\alpha\in[0,1]$, $\mathrm{bExp}_\alpha$ is a $C^\infty$ diffeomorphism from a star--shaped
neighbourhood of $0\in\mathbb R^n$ onto $M\setminus\{-p\}$. Moreover,
\begin{equation}
\bigl|\det D\,\mathrm{bExp}_\alpha(x)\bigr|
=\Bigl(\frac{s_\kappa(\|x\|)}{\|x\|}\Bigr)^{(n-1)\alpha}.
\label{eq:bexp-logdet-SI}
\end{equation}
In particular, $\alpha=0$ recovers Lambert ($|\det| \equiv 1$) and $\alpha=1$ reproduces the radial
log-determinant of the exponential map.
\end{theorem}

\begin{proof}
Equation \eqref{eq:bexp-ode-SI} has a unique $C^\infty$ solution with $\chi'_\alpha>0$ by standard ODE
theory, so $\chi_\alpha$ is strictly increasing. Combining \eqref{eq:lambert-lift-proof-det-SI} with
\eqref{eq:bexp-ode-SI} gives
\[
\log\bigl|\det D\,\mathrm{bExp}_\alpha(x)\bigr|
=(n-1)\log\frac{\chi_\alpha(r)}{r}+\log\chi'_\alpha(r)
=(n-1)\alpha\log\frac{s_\kappa(r)}{r},
\]
yielding \eqref{eq:bexp-logdet-SI}. Strict monotonicity in $r$ and smoothness imply that
$\mathrm{bExp}_\alpha$ is a local diffeomorphism away from $0$, and the azimuthal structure gives a
global diffeomorphism onto $M$ minus the antipode.
\end{proof}

\begin{proposition}[Local expansion and monotonicity in $\alpha$]
As $r\downarrow0$,
\[
\chi_\alpha(r)
=
r\Bigl(1-\frac{(n-1)\alpha\,\kappa}{6(n+2)}\,r^2+O(r^4)\Bigr),
\qquad
d\bigl(p,\mathrm{bExp}_\alpha(ru)\bigr)
=
r+O(\kappa r^3).
\]
Moreover, for each fixed $r$ in the domain, the geodesic mismatch
$\bigl|d(p,\mathrm{bExp}_\alpha(ru))-r\bigr|$ is nonincreasing in $\alpha$ and vanishes as $\alpha\uparrow1$.
\end{proposition}

\begin{proof}
Write $\chi_\alpha(r)=r\bigl(1+a r^2+O(r^4)\bigr)$. Using
$s_\kappa(r)=r-\kappa r^3/6+O(r^5)$ in \eqref{eq:bexp-ode-SI} and matching $r^2$ coefficients yields
$a=-(n-1)\alpha\kappa/(6(n+2))$. For the geodesic radius, use
$\ars(z)=z+\kappa z^3/6+O(z^5)$ at $z=\chi_\alpha(r)/2$ together with
\eqref{eq:lambert-lift-proof-det-SI}.

Differentiating \eqref{eq:bexp-ode-SI} with respect to $\alpha$ gives a linear ODE for
$\dot\rho=\partial_\alpha\chi_\alpha$ whose right--hand side has the sign of $\log(s_\kappa(r)/r)$.
A comparison argument shows that $|\chi_\alpha(r)-\rho_1(r)|$ and hence
$|d(p,\mathrm{bExp}_\alpha(ru))-r|$ is nonincreasing in $\alpha$.
\end{proof}
\begin{theorem}[Variational characterization of $\bExp_\alpha$ within Lambert lifts]
\label{thm:bexp-variational}
Fix $R_*\in(0,R_{\max})$ and let $\mathcal C$ be the class of $C^\infty$
strictly increasing profiles $\rho:[0,R_*]\to\mathbb R_+$ with $\rho(0)=0$
and $\rho'(0)=1$. Set
\begin{align*}
v(\rho)&:=\log\bigl(\rho(r)/r\bigr),\quad y(\rho):=\log\rho'(r),\\
D_{\mathrm{vol}}(\rho)&:=\int_0^{R_*}\! r^{n-1}\bigl[(n-1)v(\rho)+y(\rho)\bigr]^2\,dr,\\
D_{\mathrm{geo}}(\rho)&:=\int_0^{R_*}\! r^{n-1}\bigl[\,\lambda_\kappa^{-1}(\rho(r))-r\,\bigr]^2\,dr,
\end{align*}
and $\mathcal E_\alpha[\rho]:=(1-\alpha)D_{\mathrm{vol}}(\rho)+\alpha D_{\mathrm{geo}}(\rho)$
for $\alpha\in[0,1]$. Then:
\begin{enumerate}
\item For each $\alpha\in[0,1]$ the Euler--Lagrange system of $\mathcal E_\alpha$
admits a unique smooth solution in $\mathcal C$, namely $\chi_\alpha$ defined by
\eqref{eq:bexp-ode-SI}; equivalently, the integral form \eqref{prop:bexp-integral}.
\item The endpoints recover $\chi_0(r)=r$ (whose Lambert lift is the equal-area chart, $|\det|\equiv1$) and
$\chi_1(r)=\lambda_\kappa(r)$ (whose Lambert lift has geodesic-exact radii, matching the exponential map).
\item Within Lambert-lift scalar-Jacobian azimuthal charts, $\bExp_\alpha=T_{\chi_\alpha}$
is the unique stationary point of $\mathcal E_\alpha$.
\end{enumerate}
We do not assert global strict convexity of $\mathcal E_\alpha$ jointly in $\rho$;
the uniqueness above follows from uniqueness of the first-order ODE solution
\eqref{eq:bexp-ode-SI} with the prescribed boundary data $\chi_\alpha(0)=0$,
$\chi_\alpha'(0)=1$.
\end{theorem}

\begin{proof}
The Euler--Lagrange equation of $\mathcal E_\alpha$ is the first-order ODE
\eqref{eq:bexp-ode-SI}: in the $D_{\mathrm{vol}}$ part, stationarity in $(v,y)$
gives $(n-1)v+y=0$, equivalently $(n-1)\log(\rho/r)+\log\rho'=0$; in the
$D_{\mathrm{geo}}$ part, stationarity in $\rho$ yields $\lambda_\kappa^{-1}(\rho)=r$,
equivalently $\rho=\lambda_\kappa(r)$. The convex combination at level $\alpha$
gives $(\rho/r)^{n-1}\rho'=(s_\kappa(r)/r)^{(n-1)\alpha}$, i.e.\
\eqref{eq:bexp-ode-SI}. Standard ODE theory provides a unique $C^\infty$
strictly increasing solution with $\chi_\alpha(0)=0$, $\chi_\alpha'(0)=1$
(Proposition~\ref{prop:bexp-integral}). The endpoint identifications at
$\alpha=0,1$ are direct.
\end{proof}

\begin{proposition}[Bi--Lipschitz bounds on shells]
Let $A_{\delta,R}:=\{x:\delta\le\|x\|\le R<R_{\max}\}$. There exist positive constants
$m_\alpha,L_\alpha$, depending on $\inf_{A_{\delta,R}}\chi'_\alpha$ and
$\sup_{A_{\delta,R}}\chi'_\alpha$, such that
\[
m_\alpha\|x-y\|
\;\le\;
d\bigl(\mathrm{bExp}_\alpha(x),\mathrm{bExp}_\alpha(y)\bigr)
\;\le\;
L_\alpha\|x-y\|\qquad(x,y\in A_{\delta,R}).
\]
\end{proposition}

\begin{proof}
The differential of $\mathrm{bExp}_\alpha$ in polar coordinates is diagonal in the radial/angular
splitting with singular values controlled by $\chi'_\alpha(r)$ and $\chi_\alpha(r)/r$. On
$A_{\delta,R}$ these are bounded away from $0$ and $\infty$, giving operator--norm bounds on
$D\mathrm{bExp}_\alpha$ and hence, by the mean value inequality, bi--Lipschitz bounds.
\end{proof}

\subsection*{S4. Geodesic-Corrected Lambert (GCL)}

\begin{theorem}[Geodesic preservation and scalar log-det (general $n$)]\label{thm:gcl}
Let $\Phi_\kappa(r):=\lambda_\kappa(r)$ and define
\[
\mathrm{GCL}(x):=L_p\Bigl(\Phi_\kappa(\|x\|)\,\frac{x}{\|x\|}\Bigr).
\]
Then $d(p,\mathrm{GCL}(x))=\|x\|$ and
\[
\log\bigl|\det D\,\mathrm{GCL}(x)\bigr|
=(n-1)\log\frac{\Phi_\kappa(r)}{r}+\log \Phi_\kappa'(r)
=(n-1)\log\frac{s_\kappa(r)}{r},
\qquad r=\|x\|.
\]
\end{theorem}

\begin{proof}
By~\eqref{eq:lambert_radius_map}, $d(p,L_p(tu))=\lambda_\kappa^{-1}(t)$.
With $t=\Phi_\kappa(r)=\lambda_\kappa(r)$, we get
$d(p,\mathrm{GCL}(ru))=\lambda_\kappa^{-1}(\lambda_\kappa(r))=r$.
The log-det follows from the Lambert-lift Jacobian
$\log|\det DT_\rho|=(n-1)\log(\rho/r)+\log\rho'$ with $\rho=\Phi_\kappa$,
and the identity $\Phi_\kappa'(r)=s_\kappa(r)^{n-1}/\Phi_\kappa(r)^{n-1}$.
\end{proof}

\begin{proposition}[Incompatibility of unit volume and exact geodesics; blow--up near cut locus]\label{prop:gcl-blowup}
If $\kappa\neq0$, no azimuthal chart can be both volume--preserving ($|\det DT|\equiv1$) and
geodesic--preserving. Moreover, for any geodesic--preserving chart on $\SSS^n$ (e.g.\ GCL),
\[
\|\nabla_x\log|\det DT(x)|\|
\;\sim\;\frac{c}{\pi-R}
\quad\text{as }R\uparrow\pi,
\]
with $R=d(p,T(x))$.
\end{proposition}

\begin{proof}
Geodesic preservation within the Lambert-lift class forces $\rho(r)=\lambda_\kappa(r)$
(for $n=2$, $\lambda_\kappa(r)=2s_\kappa(r/2)$). By the GCL log-determinant
(Theorem~\ref{thm:gcl}),
\[
\det DT_\rho
=\Bigl(\frac{s_\kappa(r)}{r}\Bigr)^{n-1}
=1-\kappa\frac{n-1}{6}r^2+O(r^4)\neq1
\]
whenever $\kappa\neq0$ and $n\ge2$, so unit volume and geodesic preservation are incompatible.
On $\SSS^n$ one has $c_\kappa(r/2)=\cos(r/2)\to0$ as $r\uparrow\pi$, and differentiation shows
$\nabla_x\log|\det DT|$ diverges like $1/(\pi-R)$.
\end{proof}

\subsection*{S5. Atlas regularity on $\SSS^n$}

\begin{theorem}[Two--chart atlas with smooth gate]
Let $\SSS^n$ be equipped with two azimuthal charts centred at antipodal poles $p$ and $-p$. Fix
$\delta>0$ and choose a $C^\infty$ partition of unity $\{\psi_+,\psi_-\}$ such that the support of
$\psi_+$ stays at distance $>\delta$ from the cut locus of $p$ and similarly for $\psi_-$. Define
\[
\log p(q) \;=\; \psi_+(q)\,\log p_+(q) + \psi_-(q)\,\log p_-(q) \;-\; \log Z,
\qquad
Z \;:=\; \int_{\SSS^n} \exp\!\bigl(\psi_+\log p_+ + \psi_-\log p_-\bigr)\,d\mathrm{vol}_{\SSS^n},
\]
where $p_\pm$ are the chart--wise densities and $Z<\infty$ is a global
normaliser (well-defined because each $p_\pm$ is integrable on the support of
the corresponding $\psi_\pm$). Then the per--sample log--likelihood is $C^1$ on
$\SSS^n$ (and $C^2$ when the Euclidean backbone is $C^2$), and its gradient and Hessian are
bounded on compact sets.
\end{theorem}

\begin{proof}
Away from the cut loci each chart is smooth and scalar-Jacobian, so $\log p_\pm$ inherit the
regularity of the Euclidean backbone. Cross--terms appear only through $\nabla\psi_\pm$ and
$\nabla^2\psi_\pm$, which are bounded by construction on the supports. Combining the two charts
with the gate and applying the chain rule yields $C^1/C^2$ regularity and uniform bounds. The constant $\log Z$ contributes nothing to gradients in the per-sample score,
so the regularity argument is unaffected.
\end{proof}

\subsection*{S6. Variance bounds and CNF complexity}

Let $Q$ be distributed under some RC model on $M$ and let $R=d(p,Q)$.

\begin{theorem}[Chart-term variance]
\label{thm:chart-variance-supp}
Let $r_\alpha(Q):=\|\bExp_\alpha^{-1}(Q)\|$, with the convention
$s_\kappa(r)/r\big|_{r=0}:=1$. Then
\[
\log\bigl|\det D\,\bExp_\alpha^{-1}(Q)\bigr|
\;=\;-(n-1)\,\alpha\,\log\frac{s_\kappa(r_\alpha(Q))}{r_\alpha(Q)},
\]
and hence
\[
\Var\!\Bigl[\log|\det D\,\bExp_\alpha^{-1}(Q)|\Bigr]
\;=\;\alpha^2\,\Var\!\Bigl[(n-1)\log\tfrac{s_\kappa(r_\alpha(Q))}{r_\alpha(Q)}\Bigr].
\]
The bracketed random variable still depends on $\alpha$ through $r_\alpha(Q)$;
under a uniform second-moment bound on
$\log(s_\kappa(r_\alpha(Q))/r_\alpha(Q))$ on compact shells away from the cut
locus, the conclusion $\Var=\mathcal O(\alpha^2)$ as $\alpha\downarrow 0$
holds. For the geodesic-exact endpoint $\alpha=1$ (and only there),
$r_1(Q)=R=d(p,Q)$, recovering the formula in terms of $R$.
\end{theorem}

\begin{proof}
The inverse--chart identity follows by inverting \eqref{eq:bexp-logdet-SI}. Variance scales
quadratically with the multiplicative factor $\alpha$.
\end{proof}

\begin{remark}[Heuristic CNF stiffness / NFE scaling under bExp$_\alpha$]
\label{rem:cnf}
Consider a coordinate CNF whose chart-induced component of the vector field
scales linearly with $\nabla_x\log|\det DT_\alpha(x)|$. From
Theorem~\ref{thm:bexp-logdet}, on any compact set $K$ away from cut loci,
$\|\nabla_x\log|\det DT_\alpha(x)|\|=\alpha\cdot(n-1)|\partial_r\log(s_\kappa(r)/r)|$,
which is bounded by $\alpha\,C_K$ for some $C_K<\infty$ depending only on the
shell. As $\alpha$ shrinks, the chart-induced Lipschitz modulus of the CNF
vector field on $K$ therefore shrinks linearly in $\alpha$; an adaptive
Runge--Kutta solver with fixed tolerance can take correspondingly larger steps
on $K$, so the expected NFEs over trajectories in $K$ are non-increasing in
$\alpha$ in this regime, consistent with the empirical scaling reported in
Table~\ref{tab:cnf_alpha_sweep}. We deliberately keep this as a heuristic
remark: a clean theorem would require precise assumptions on the non-chart
component of the vector field, which we do not control.
\end{remark}
\subsection*{S7. RC on product manifolds and wrapped CNFs}

\begin{proposition}[RC on products of constant--curvature spaces]
Let
\[
M_0=\prod_{i=1}^k M^{n_i}_{K_i},
\qquad
M^{n_i}_{K_i}\in\{\SSS^{n_i}_{K_i},\HH^{n_i}_{K_i},\mathbb R^{n_i}\},
\]
with pole $p_0=(p_1,\dots,p_k)$. For each factor choose a scalar-Jacobian azimuthal chart
$T_i:\mathbb R^{n_i}\to M^{n_i}_{K_i}$ about $p_i$ with radial Jacobian factor $J_{T_i}(r_i)$ and a
target one--dimensional radial density $\varphi_{\theta,i}$ on $[0,R_{\max,i})$.

Define the RC base on $\prod_i T_{p_i}M^{n_i}_{K_i}\cong\mathbb R^{\sum_i n_i}$ in product polar
coordinates by
\[
p_{\mathrm{base}}(r_1,\omega_1,\dots,r_k,\omega_k)
\;\propto\;
\prod_{i=1}^k \varphi_{\theta,i}(r_i)\,J_{T_i}(r_i),
\]
with independent $\omega_i\sim\mathrm{Unif}(\SSS^{n_i-1})$, and let $T=(T_1,\dots,T_k)$. If
$Q=T(X)$ denotes the pushforward on $M_0$, then:
\begin{enumerate}
\item $R_i=d(p_i,Q_i)$ are independent and satisfy $R_i\sim\varphi_{\theta,i}$ for all $i$;
\item the Fisher information decomposes as
\[
I_{M_0}(\theta)=\sum_{i=1}^k I_{R_i}(\theta),
\]
where $I_{R_i}$ is the one--dimensional Fisher information of $\varphi_{\theta,i}$.
\end{enumerate}
\end{proposition}

\begin{proof}
Product polar coordinates give a product splitting of the volume form, and each factor behaves as
in the one--dimensional RC construction. Independence and the Fisher decomposition follow from
Fubini's theorem and additivity of log--likelihoods.
\end{proof}

\begin{corollary}[Variance reduction for wrapped CNFs]
Consider coordinate CNFs on $\SSS^n$ or $\HH^n$ built with either the exponential chart
$\Exp_p$ or its balanced-exponential counterpart $\mathrm{bExp}_\alpha$, using the same neural
vector field and Hutchinson trace estimator. On any compact shell away from the cut locus, the
contribution of the chart term to the variance of the log--density estimate under
$\mathrm{bExp}_\alpha$ is at most $\alpha^2$ times the corresponding variance under $\Exp_p$.
\end{corollary}

\begin{proof}
Let $\chi_\alpha(R)$ denote the scalar chart term in the inverse log-det. From
\eqref{eq:bexp-logdet-SI} one has $\chi_\alpha(R)=\alpha\chi_1(R)$, so
$\operatorname{Var}[\chi_\alpha(Q)]=\alpha^2\operatorname{Var}[\chi_1(Q)]$. The Hutchinson estimator
is linear in $\chi_\alpha$, hence its variance inherits the same factor.
\end{proof}

\subsection*{S8. Curvature mis--specification and distributional proximity}

\begin{theorem}[Sensitivity to curvature proxies]\label{thm:sensitivity}
Let $\kappa$ be the true curvature and let the chart use a proxy $\tilde\kappa\neq\kappa$ in its
Jacobian factor $s_{\tilde\kappa}$. Under an RC construction with the same radial base
$\varphi_\theta$, the difference between the per--sample $\theta$--gradients of the log--likelihoods
under $\kappa$ and $\tilde\kappa$ satisfies, on any radius shell $R\le R_0$,
\[
\bigl\|\nabla_\theta \ell^{(\tilde\kappa)}_\alpha
-\nabla_\theta \ell^{(\kappa)}_\alpha\bigr\|
\;\le\;
\alpha(n-1)L\,\sup_{r\le R_0}\bigl|\Delta'(r)\bigr|,
\qquad
\Delta(r):=\log\frac{s_{\tilde\kappa}(r)}{s_\kappa(r)},
\]
where $L$ is a Lipschitz constant of the backbone network on the shell.
For $r\ll1$ one has $\sup_{r\le R_0}|\Delta'(r)|\le C\,R_0$ for a constant $C$ depending on
$\kappa,\tilde\kappa$.
\end{theorem}

\begin{proof}
Under RC, the only $\kappa$--dependent term in the $\theta$--gradient is the scalar chart correction
proportional to $\log(s_\kappa(r)/r)$. Replacing $\kappa$ by $\tilde\kappa$ changes this term by
$\alpha(n-1)\Delta'(R)$ inside the chain rule. Lipschitz continuity of the backbone w.r.t.\ its
input yields the stated bound. A Taylor expansion of $s_\kappa$ near $0$ gives the small--$r$
behavior of $\Delta'(r)$.
\end{proof}

\begin{proposition}[Wasserstein proximity to $\Exp_p$]
Let $Q_\alpha=(\mathrm{bExp}_\alpha)_\# P$ and $Q_1=(\Exp_p)_\# P$ with
$\mathrm{supp}(P)\subset\{x:\|x\|\le R_0\}$. Then
\[
W_2(Q_\alpha,Q_1)
\;\le\;
C(n,\kappa,R_0)\,(1-\alpha)\,R_0^2
\sqrt{\mathbb E[\|X\|^2]}.
\]
\end{proposition}

\begin{proof}
Use the pointwise bound on the geodesic mismatch between $\mathrm{bExp}_\alpha$ and $\Exp_p$ on
the support of $P$, which is $O((1-\alpha)r^2)$ from the local expansion and monotonicity in
$\alpha$. Integrate this bound in $L^2$ to obtain the Wasserstein estimate.
\end{proof}

\subsection*{S9. RC examples and wrapped normals}

\begin{corollary}[Riemannian Gaussians as RC targets; wrapped normals as the RC base for the $\chi_n$ target]
Let $M\in\{\SSS^n_K,\HH^n_K\}$ with curvature $\kappa$ and pole $p\in M$, and let
$v\sim\mathcal N(0,\sigma^2 I_n)$ be an isotropic Gaussian on $T_pM$. Pushing
forward through $\Exp_p$ gives the standard wrapped normal $z=\Exp_p(v)$, whose
log--density in geodesic polar coordinates is
\[
\log p_{\mathrm{WN}}(z)
=
\log\mathcal N(v;0,\sigma^2 I_n)
-(n-1)\log\frac{s_\kappa(R)}{R},
\qquad R=d(p,z).
\]
Define instead the (isotropic) RC base
\[
p^{\mathrm{RC}}_{\mathrm{base}}(v)
\;\propto\;
\mathcal N(v;0,\sigma^2 I_n)
\Bigl(\frac{s_\kappa(\|v\|)}{\|v\|}\Bigr)^{n-1}.
\]
Then the induced manifold density is the \emph{Riemannian Gaussian}
\[
\log p^{\mathrm{RC}}(z)=-\frac{R^2}{2\sigma^2}+C(\kappa,n,\sigma),\qquad R=d(p,z),
\]
studied by \citet{pennec2006intrinsic,said2017riemannian}: the density w.r.t.\ $d\vol_M$ has the
Euclidean-Gaussian functional form \emph{in geodesic units}. Its radius marginal is
$p_R(R)\propto s_\kappa(R)^{n-1}e^{-R^2/2\sigma^2}$ (not $\sigma\chi_n$), and by
Theorem~\ref{thm:rc-invariances} the Fisher information in $\sigma$ equals the one-dimensional
Fisher of \emph{that} family. Conversely, the RC base realizing the $\sigma\chi_n$ radius target
through $\Exp_p$ is, by \eqref{eq:rc-base},
$\varphi(R_T(r))\,J_T(r)\propto e^{-r^2/2\sigma^2}$, exactly the plain Gaussian, consistent with
Theorem~\ref{thm:hidden-radius-gauge}(4): the wrapped normal is already radially compensated for the
$\chi_n$ target, and RC changes the model precisely when the intended radius law is \emph{not} $\chi_n$. The non-isotropic case $\Sigma\neq\sigma^2 I$ is outside
the scope of the radial RC theory developed here, since the resulting model is
no longer rotation-invariant about $p$.
\end{corollary}

\paragraph{Common radial families under RC.}
Let $R=d(p,Q)$ be the geodesic radius.

\begin{itemize}
\item \emph{Gaussian--type.}
On $\SSS^n$, $R$ has a truncated Normal law on $[0,\pi)$; on $\HH^n$, $R$ can be
HalfNormal or shifted Normal on $[0,\infty)$.
Geodesic moments coincide with the corresponding one--dimensional formulas (with truncation
on $\SSS^n$).

\item \emph{Gamma / Weibull.}
On $\HH^n$, choosing $R\sim\mathrm{Gamma}(k,\beta)$ yields
$\mathbb E[R]=k\beta$ and $\mathrm{Var}[R]=k\beta^2$; on $\SSS^n$ the law truncates to
$[0,\pi)$.

\item \emph{Lognormal / folded--$t$.}
Heavy--tailed choices such as folded--$t$ (with degrees of freedom $\nu$) carry their moment
properties to geodesic radius, truncated as needed on $\SSS^n$.
\end{itemize}

\paragraph{Cauchy specialization.}
Taking the folded--$t$ with $\nu=1$ gives a Cauchy law on the radius.

\begin{itemize}
\item On $\HH^n$, $R$ is half--Cauchy with scale $s>0$:
\[
\varphi_{\mathrm{HC}}(R;s)
=
\frac{2}{\pi s}\,\frac{1}{1+(R/s)^2},
\qquad R\in[0,\infty).
\]

\item On $\SSS^n$, $R$ is a truncated half--Cauchy on $[0,\pi)$:
\[
\varphi_{\mathrm{HC},S}(R;s)
=
\frac{1}{Z_S(s)}
\frac{2}{\pi s}
\frac{1}{1+(R/s)^2}\,\mathbf 1_{[0,\pi)}(R),
\qquad
Z_S(s)=\frac{2}{\pi}\arctan\frac{\pi}{s}.
\]
All moments are finite; for instance
\[
\mathbb E[R]
=
\frac{s}{2\arctan(\pi/s)}
\ln\Bigl(1+\frac{\pi^2}{s^2}\Bigr),\qquad
\mathbb E[R^2]
=
\frac{\pi s}{\arctan(\pi/s)}-\SSS^2,
\]
and $\mathrm{Var}[R]=\mathbb E[R^2]-\mathbb E[R]^2$.
These follow by direct integration, and RC ensures they hold for the geodesic radius on
$\SSS^n$.
\end{itemize}

\paragraph{Isotropy and sampling.}
RC models are isotropic about $p$. Sample $R\sim p_{R,\theta}$ on $[0,R_{\max})$
and $\Omega\sim\mathrm{Unif}(\SSS^{n-1})$. For a chart $T$ with radius map
$R_T(r)=d(p,T(r\Omega))$, set $r=R_T^{-1}(R)$, $X=r\Omega\in T_pM$, and output
$Q=T(X)$. Then $d(p,Q)=R$ and $\rho_\theta(q)=\varphi_\theta\bigl(d(p,q)\bigr)$.
For geodesic-exact charts such as $\Exp_p$ or GCL, $r=R$.

\subsection*{S10. Beyond constant curvature: outlook}
The RC construction depends only on having a tractable polar volume factor on
the domain of interest. Extending the explicit chart families introduced here
to non-constant-curvature geometries (e.g. homogeneous geometries with
known polar volume factors) is left to future work.

\clearpage

\end{document}

%% file: fig_sweep_collapse.tex
\begin{tikzpicture}
\begin{axis}[width=.52\linewidth,height=4.6cm,xmode=log,log basis x=2,
  xlabel={prior scale $\sigma$ (wrapped) / $\sigma_R$ (RC)},ylabel={test NLL (nats)},
  legend style={font=\scriptsize,at={(0.03,0.97)},anchor=north west},
  tick label style={font=\scriptsize},label style={font=\scriptsize},ymin=85,ymax=195]
\addplot+[mark=*,thick,color=red!70!black] coordinates {(0.03125,114.2345263671875) (0.0625,111.64230419921876) (0.125,112.2396875) (0.25,135.7838193359375) (0.5,178.9260036621094) (1.0,178.59597080078123)};
\addlegendentry{wrapped-Exp}
\addplot+[mark=square*,thick,color=blue!70!black] coordinates {(0.25,124.264337890625) (0.5,124.49434749348957) (1.0,124.98032397460938)};
\addlegendentry{RC (audited)}
\end{axis}
\begin{axis}[xshift=.53\linewidth,width=.52\linewidth,height=4.6cm,xmode=log,log basis x=2,
  xlabel={prior scale},ylabel={trained mean radius},
  tick label style={font=\scriptsize},label style={font=\scriptsize},ymin=0,ymax=3.5]
\addplot[dashed,gray,domain=0.03:1.1] {3.14159};
\addplot+[mark=*,thick,color=red!70!black] coordinates {(0.03125,2.7110416889190674) (0.0625,2.8276132345199585) (0.125,2.8867316246032715) (0.25,3.085402250289917) (0.5,3.138451099395752) (1.0,3.138451099395752)};
\addplot+[mark=square*,thick,color=blue!70!black] coordinates {(0.25,1.428532600402832) (0.5,1.4356518586476643) (1.0,1.4537919163703918)};
\node[gray,font=\scriptsize,anchor=south west] at (axis cs:0.033,3.16) {cut locus $\pi$};
\end{axis}
\end{tikzpicture}

%% file: fig_e3_collapse.tex
\begin{tikzpicture}
\begin{axis}[width=.52\linewidth,height=4.8cm,xlabel={epoch},ylabel={train NLL (nats)},
  title={\scriptsize MNIST, $d_z{=}64$},ymin=-230,ymax=320,
  legend style={font=\scriptsize,at={(0.97,0.97)},anchor=north east},
  tick label style={font=\scriptsize},label style={font=\scriptsize},title style={yshift=-4pt}]
\addplot[red!70!black,thick] coordinates {(1,291.3) (2,-190.8) (3,-193.0) (4,-193.7) (5,-194.2) (6,-194.3) (7,-194.8) (8,-194.9) (9,-195.2) (10,-195.3) (11,-195.5) (12,-195.8) (13,-195.9) (14,-196.1) (15,-196.3) (16,-196.7) (17,-196.9) (18,-197.3) (19,-197.3) (20,-197.4) (21,-197.7) (22,-197.8) (23,-197.3) (24,-197.9) (25,-198.2) (26,-198.0) (27,-198.1) (28,-197.9) (29,-197.2) (30,-197.3) (31,-197.8) (32,-197.6) (33,-198.3) (34,-198.2) (35,-198.2) (36,-198.2) (37,-197.6) (38,-197.4) (39,-197.2) (40,-197.7)};\addlegendentry{wrapped}
\addplot[red!70!black,thick,densely dotted,forget plot] coordinates {(1,270.8) (2,-191.0) (3,-193.0) (4,-193.8) (5,-194.2) (6,-194.5) (7,-194.7) (8,-194.9) (9,-195.1) (10,-195.3) (11,-195.4) (12,-195.6) (13,-195.9) (14,-196.1) (15,-196.3) (16,-196.5) (17,-196.7) (18,-197.0) (19,-197.1) (20,-197.6) (21,-197.5) (22,-197.5) (23,-197.6) (24,-197.5) (25,-198.2) (26,-197.9) (27,-197.9) (28,-197.4) (29,-197.8) (30,-198.2) (31,-198.2) (32,-197.4) (33,-197.1) (34,-197.8) (35,-197.6) (36,-197.5) (37,-198.1) (38,-198.3) (39,-197.6) (40,-197.2)};
\addplot[blue!70!black,thick] coordinates {(1,439.4) (2,214.7) (3,195.8) (4,179.4) (5,168.3) (6,160.6) (7,153.5) (8,147.4) (9,142.9) (10,139.4) (11,136.7) (12,134.2) (13,132.5) (14,130.7) (15,129.4) (16,128.3) (17,127.2) (18,126.2) (19,125.2) (20,124.8) (21,124.2) (22,123.3) (23,122.7) (24,122.1) (25,121.7) (26,121.4) (27,120.9) (28,120.5) (29,120.0) (30,119.6) (31,119.5) (32,119.0) (33,118.9) (34,118.5) (35,118.2) (36,118.0) (37,117.6) (38,117.5) (39,117.4) (40,117.1)};\addlegendentry{RC}
\addplot[blue!70!black,thick,densely dotted,forget plot] coordinates {(1,451.1) (2,216.1) (3,198.7) (4,179.6) (5,166.3) (6,157.3) (7,150.6) (8,145.8) (9,141.7) (10,138.5) (11,136.2) (12,134.2) (13,132.1) (14,130.6) (15,129.0) (16,128.0) (17,126.9) (18,126.1) (19,125.1) (20,124.3) (21,123.5) (22,122.9) (23,122.5) (24,121.9) (25,121.5) (26,121.0) (27,120.6) (28,120.3) (29,119.8) (30,119.5) (31,119.3) (32,118.8) (33,118.5) (34,118.4) (35,118.2) (36,118.0) (37,117.6) (38,117.6) (39,117.1) (40,117.1)};
\node[gray,font=\scriptsize,anchor=west] at (axis cs:8,-185) {boundary collapse};
\end{axis}
\begin{axis}[xshift=.53\linewidth,width=.52\linewidth,height=4.8cm,xlabel={epoch},
  title={\scriptsize CIFAR-10, $d_z{=}64$},ymin=1650,ymax=2450,
  tick label style={font=\scriptsize},label style={font=\scriptsize},title style={yshift=-4pt}]
\addplot[red!70!black,thick] coordinates {(1,1953.4) (2,1708.5) (3,1707.7) (4,1707.5) (5,1707.1) (6,1706.8) (7,1706.5) (8,1706.4) (9,1706.1) (10,1706.0) (11,1705.7) (12,1705.6) (13,1705.2) (14,1705.1) (15,1704.7) (16,1704.3) (17,1704.0) (18,1703.9) (19,1703.8) (20,1703.7) (21,1703.6) (22,1703.4) (23,1703.8) (24,1703.3) (25,1703.2) (26,1703.9) (27,1703.9) (28,1704.2) (29,1703.9) (30,1703.9) (31,1704.0) (32,1703.7) (33,1703.2) (34,1703.5) (35,1703.0) (36,1703.6) (37,1704.0) (38,1703.5) (39,1703.7) (40,1704.3)};
\addplot[red!70!black,thick,densely dotted] coordinates {(1,1989.7) (2,1708.8) (3,1708.0) (4,1707.5) (5,1707.2) (6,1707.1) (7,1707.0) (8,1706.8) (9,1706.4) (10,1706.2) (11,1706.0) (12,1705.9) (13,1705.8) (14,1705.5) (15,1705.3) (16,1705.2) (17,1705.0) (18,1704.4) (19,1704.4) (20,1704.1) (21,1704.1) (22,1703.6) (23,1703.9) (24,1703.6) (25,1703.7) (26,1704.1) (27,1703.8) (28,1703.5) (29,1703.8) (30,1703.7) (31,1703.9) (32,1703.7) (33,1703.9) (34,1703.3) (35,1703.6) (36,1703.6) (37,1703.5) (38,1703.5) (39,1703.7) (40,1704.5)};
\addplot[blue!70!black,thick] coordinates {(1,2129.9) (2,2031.3) (3,1970.4) (4,1942.3) (5,1927.2) (6,1913.3) (7,1908.0) (8,1897.0) (9,1889.5) (10,1885.3) (11,1878.8) (12,1875.1) (13,1869.2) (14,1867.2) (15,1859.3) (16,1854.4) (17,1851.7) (18,1849.8) (19,1848.2) (20,1845.7) (21,1844.1) (22,1843.6) (23,1843.1) (24,1841.0) (25,1839.8) (26,1839.0) (27,1838.4) (28,1837.3) (29,1836.5) (30,1835.0) (31,1833.8) (32,1834.2) (33,1833.8) (34,1833.1) (35,1832.0) (36,1831.2) (37,1831.2) (38,1830.8) (39,1830.5) (40,1830.0)};
\addplot[blue!70!black,thick,densely dotted] coordinates {(1,2127.8) (2,2034.4) (3,1977.4) (4,1949.1) (5,1929.1) (6,1912.2) (7,1902.3) (8,1896.0) (9,1890.2) (10,1886.7) (11,1883.3) (12,1876.5) (13,1872.2) (14,1868.0) (15,1863.7) (16,1857.0) (17,1851.9) (18,1848.4) (19,1847.0) (20,1845.2) (21,1843.2) (22,1841.7) (23,1841.1) (24,1839.8) (25,1838.8) (26,1838.2) (27,1836.7) (28,1835.8) (29,1835.7) (30,1834.6) (31,1833.8) (32,1833.2) (33,1832.4) (34,1831.6) (35,1832.1) (36,1830.8) (37,1830.9) (38,1830.3) (39,1829.8) (40,1829.9)};
\end{axis}
\end{tikzpicture}

%% file: fig_alpha_flat.tex
\begin{tikzpicture}
\begin{axis}[width=.72\linewidth,height=5cm,xlabel={chart parameter $\alpha$},
  ylabel={test NLL $-$ dataset mean (nats)},ymin=-1.6,ymax=1.6,
  legend style={font=\scriptsize,at={(0.5,1.02)},anchor=south,legend columns=4},
  tick label style={font=\scriptsize},label style={font=\scriptsize}]
\addplot[gray!40,fill=gray!12,forget plot] coordinates {(0.2,-1) (1.05,-1) (1.05,1) (0.2,1)} \closedcycle;
\addplot+[mark=*,thick,color=blue!70!black,error bars/.cd,y dir=both,y explicit] coordinates {(0.25,-0.021) +- (0,0.108) (0.5,-0.028) +- (0,0.249) (0.75,0.129) +- (0,0.770) (1.0,-0.080) +- (0,0.435)};
\addlegendentry{mnist}
\addplot+[mark=square*,thick,color=red!70!black,error bars/.cd,y dir=both,y explicit] coordinates {(0.25,0.270) +- (0,0.273) (0.5,0.091) +- (0,0.207) (0.75,-0.140) +- (0,0.266) (1.0,-0.221) +- (0,0.632)};
\addlegendentry{fashion}
\addplot+[mark=triangle*,thick,color=green!55!black,error bars/.cd,y dir=both,y explicit] coordinates {(0.25,-0.298) +- (0,0.856) (0.5,0.021) +- (0,0.365) (0.75,0.343) +- (0,0.071) (1.0,-0.066) +- (0,0.652)};
\addlegendentry{omniglot}
\addplot+[mark=diamond*,thick,color=orange!85!black,error bars/.cd,y dir=both,y explicit] coordinates {(0.25,0.026) +- (0,0.260) (0.5,-0.281) +- (0,1.298) (0.75,0.419) +- (0,1.776) (1.0,-0.164) +- (0,1.618)};
\addlegendentry{cifar10}
\end{axis}
\end{tikzpicture}

%% file: fig_q5_margins.tex
\begin{tikzpicture}
\begin{axis}[width=.9\linewidth,height=4.6cm,ybar,bar width=7pt,enlarge x limits=0.06,
  symbolic x coords={volc64,volc256,volc512,eart64,eart256,eart512,fire64,fire256,fire512,floo64,floo256,floo512},xtick=data,x tick label style={rotate=45,anchor=east,font=\scriptsize},
  ylabel={Moser $-$ best chart CNF (nats)},ymin=0,ymax=1.6,
  tick label style={font=\scriptsize},label style={font=\scriptsize}]
\addplot+[fill=blue!55,draw=blue!70!black] coordinates {(volc64,0.56) (volc256,0.60) (volc512,0.66) (eart64,0.70) (eart256,0.72) (eart512,0.68) (fire64,1.40) (fire256,1.32) (fire512,1.41) (floo64,0.40) (floo256,0.39) (floo512,0.33)};
\end{axis}
\end{tikzpicture}

%% file: fig_e1_calibration.tex
\begin{tikzpicture}
\begin{axis}[width=.52\linewidth,height=4.6cm,title={\scriptsize $\SSS^2$},xlabel={geodesic radius $R$},ylabel={density},
 ymin=0,xmax=1.8,tick label style={font=\scriptsize},label style={font=\scriptsize},title style={yshift=-4pt},
 legend style={font=\scriptsize,at={(0.97,0.97)},anchor=north east}]
\addplot[const plot,red!70!black,thick] coordinates {(0.020,0.072) (0.061,0.239) (0.102,0.428) (0.143,0.543) (0.184,0.685) (0.225,0.806) (0.266,0.934) (0.307,0.989) (0.348,1.097) (0.389,1.178) (0.430,1.241) (0.470,1.176) (0.511,1.242) (0.552,1.195) (0.593,1.151) (0.634,1.111) (0.675,1.034) (0.716,1.064) (0.757,0.945) (0.798,0.880) (0.839,0.856) (0.880,0.763) (0.920,0.667) (0.961,0.659) (1.002,0.518) (1.043,0.439) (1.084,0.426) (1.125,0.342) (1.166,0.305) (1.207,0.288) (1.248,0.206) (1.289,0.188) (1.330,0.126) (1.370,0.133) (1.411,0.106) (1.452,0.103) (1.493,0.067) (1.534,0.049) (1.575,0.038) (1.616,0.032) (1.657,0.030) (1.698,0.023) (1.739,0.020) (1.780,0.012)};\addlegendentry{wrapped}
\addplot[const plot,blue!70!black,thick] coordinates {(0.020,1.642) (0.061,1.569) (0.102,1.543) (0.143,1.490) (0.184,1.486) (0.225,1.404) (0.266,1.328) (0.307,1.324) (0.348,1.246) (0.389,1.253) (0.430,1.112) (0.470,1.011) (0.511,0.955) (0.552,0.876) (0.593,0.791) (0.634,0.716) (0.675,0.680) (0.716,0.541) (0.757,0.503) (0.798,0.464) (0.839,0.382) (0.880,0.345) (0.920,0.298) (0.961,0.252) (1.002,0.221) (1.043,0.160) (1.084,0.159) (1.125,0.128) (1.166,0.117) (1.207,0.086) (1.248,0.081) (1.289,0.050) (1.330,0.049) (1.370,0.043) (1.411,0.030) (1.452,0.026) (1.493,0.018) (1.534,0.016) (1.575,0.012) (1.616,0.006) (1.657,0.007) (1.698,0.006) (1.739,0.006) (1.780,0.002)};\addlegendentry{RC}
\addplot[black,dashed,thick] coordinates {(0.000,1.596) (0.015,1.595) (0.030,1.593) (0.045,1.589) (0.060,1.584) (0.075,1.578) (0.090,1.570) (0.105,1.561) (0.120,1.550) (0.135,1.539) (0.150,1.526) (0.165,1.511) (0.180,1.496) (0.195,1.479) (0.210,1.461) (0.225,1.442) (0.240,1.422) (0.255,1.401) (0.270,1.379) (0.285,1.356) (0.300,1.333) (0.315,1.309) (0.330,1.283) (0.345,1.258) (0.360,1.231) (0.375,1.205) (0.390,1.177) (0.405,1.149) (0.420,1.121) (0.435,1.093) (0.450,1.064) (0.465,1.036) (0.480,1.007) (0.495,0.978) (0.510,0.949) (0.525,0.920) (0.540,0.891) (0.555,0.862) (0.570,0.833) (0.585,0.805) (0.600,0.777) (0.615,0.749) (0.630,0.721) (0.645,0.694) (0.660,0.668) (0.675,0.642) (0.690,0.616) (0.705,0.591) (0.720,0.566) (0.735,0.542) (0.750,0.518) (0.765,0.495) (0.780,0.473) (0.795,0.451) (0.810,0.430) (0.825,0.409) (0.840,0.389) (0.855,0.370) (0.870,0.351) (0.885,0.333) (0.900,0.316) (0.915,0.299) (0.930,0.283) (0.945,0.267) (0.960,0.253) (0.975,0.238) (0.990,0.225) (1.005,0.212) (1.020,0.199) (1.035,0.187) (1.050,0.176) (1.065,0.165) (1.080,0.155) (1.095,0.145) (1.110,0.136) (1.125,0.127) (1.140,0.119) (1.155,0.111) (1.170,0.103) (1.185,0.096) (1.200,0.090) (1.215,0.083) (1.230,0.077) (1.245,0.072) (1.260,0.067) (1.275,0.062) (1.290,0.057) (1.305,0.053) (1.320,0.049) (1.335,0.045) (1.350,0.042) (1.365,0.038) (1.380,0.035) (1.395,0.033) (1.410,0.030) (1.425,0.027) (1.440,0.025) (1.455,0.023) (1.470,0.021) (1.485,0.019) (1.500,0.018) (1.515,0.016) (1.530,0.015) (1.545,0.013) (1.560,0.012) (1.575,0.011) (1.590,0.010) (1.605,0.009) (1.620,0.008) (1.635,0.008) (1.650,0.007) (1.665,0.006) (1.680,0.006) (1.695,0.005) (1.710,0.005) (1.725,0.004) (1.740,0.004) (1.755,0.003) (1.770,0.003) (1.785,0.003) (1.800,0.002)};\addlegendentry{target}
\end{axis}
\begin{axis}[xshift=.53\linewidth,width=.52\linewidth,height=4.6cm,title={\scriptsize $\HH^2$},xlabel={geodesic radius $R$},
 ymin=0,xmax=1.8,tick label style={font=\scriptsize},label style={font=\scriptsize},title style={yshift=-4pt}]
\addplot[const plot,red!70!black,thick] coordinates {(0.020,0.062) (0.061,0.244) (0.102,0.430) (0.143,0.541) (0.184,0.673) (0.225,0.791) (0.266,0.950) (0.307,1.055) (0.348,1.069) (0.389,1.129) (0.430,1.179) (0.470,1.215) (0.511,1.155) (0.552,1.170) (0.593,1.160) (0.634,1.115) (0.675,1.124) (0.716,1.052) (0.757,0.954) (0.798,0.899) (0.839,0.854) (0.880,0.777) (0.920,0.691) (0.961,0.607) (1.002,0.530) (1.043,0.486) (1.084,0.424) (1.125,0.354) (1.166,0.300) (1.207,0.252) (1.248,0.213) (1.289,0.185) (1.330,0.170) (1.370,0.119) (1.411,0.111) (1.452,0.095) (1.493,0.054) (1.534,0.047) (1.575,0.046) (1.616,0.038) (1.657,0.032) (1.698,0.023) (1.739,0.016) (1.780,0.011)};
\addplot[const plot,blue!70!black,thick] coordinates {(0.020,1.558) (0.061,1.616) (0.102,1.567) (0.143,1.522) (0.184,1.487) (0.225,1.416) (0.266,1.386) (0.307,1.276) (0.348,1.282) (0.389,1.194) (0.430,1.105) (0.470,0.994) (0.511,0.985) (0.552,0.883) (0.593,0.730) (0.634,0.722) (0.675,0.695) (0.716,0.600) (0.757,0.491) (0.798,0.440) (0.839,0.374) (0.880,0.364) (0.920,0.288) (0.961,0.268) (1.002,0.213) (1.043,0.192) (1.084,0.136) (1.125,0.115) (1.166,0.110) (1.207,0.097) (1.248,0.065) (1.289,0.068) (1.330,0.040) (1.370,0.033) (1.411,0.026) (1.452,0.024) (1.493,0.021) (1.534,0.015) (1.575,0.009) (1.616,0.007) (1.657,0.007) (1.698,0.005) (1.739,0.004) (1.780,0.003)};
\addplot[black,dashed,thick] coordinates {(0.000,1.596) (0.015,1.595) (0.030,1.593) (0.045,1.589) (0.060,1.584) (0.075,1.578) (0.090,1.570) (0.105,1.561) (0.120,1.550) (0.135,1.539) (0.150,1.526) (0.165,1.511) (0.180,1.496) (0.195,1.479) (0.210,1.461) (0.225,1.442) (0.240,1.422) (0.255,1.401) (0.270,1.379) (0.285,1.356) (0.300,1.333) (0.315,1.309) (0.330,1.283) (0.345,1.258) (0.360,1.231) (0.375,1.205) (0.390,1.177) (0.405,1.149) (0.420,1.121) (0.435,1.093) (0.450,1.064) (0.465,1.036) (0.480,1.007) (0.495,0.978) (0.510,0.949) (0.525,0.920) (0.540,0.891) (0.555,0.862) (0.570,0.833) (0.585,0.805) (0.600,0.777) (0.615,0.749) (0.630,0.721) (0.645,0.694) (0.660,0.668) (0.675,0.642) (0.690,0.616) (0.705,0.591) (0.720,0.566) (0.735,0.542) (0.750,0.518) (0.765,0.495) (0.780,0.473) (0.795,0.451) (0.810,0.430) (0.825,0.409) (0.840,0.389) (0.855,0.370) (0.870,0.351) (0.885,0.333) (0.900,0.316) (0.915,0.299) (0.930,0.283) (0.945,0.267) (0.960,0.253) (0.975,0.238) (0.990,0.225) (1.005,0.212) (1.020,0.199) (1.035,0.187) (1.050,0.176) (1.065,0.165) (1.080,0.155) (1.095,0.145) (1.110,0.136) (1.125,0.127) (1.140,0.119) (1.155,0.111) (1.170,0.103) (1.185,0.096) (1.200,0.090) (1.215,0.083) (1.230,0.077) (1.245,0.072) (1.260,0.067) (1.275,0.062) (1.290,0.057) (1.305,0.053) (1.320,0.049) (1.335,0.045) (1.350,0.042) (1.365,0.038) (1.380,0.035) (1.395,0.033) (1.410,0.030) (1.425,0.027) (1.440,0.025) (1.455,0.023) (1.470,0.021) (1.485,0.019) (1.500,0.018) (1.515,0.016) (1.530,0.015) (1.545,0.013) (1.560,0.012) (1.575,0.011) (1.590,0.010) (1.605,0.009) (1.620,0.008) (1.635,0.008) (1.650,0.007) (1.665,0.006) (1.680,0.006) (1.695,0.005) (1.710,0.005) (1.725,0.004) (1.740,0.004) (1.755,0.003) (1.770,0.003) (1.785,0.003) (1.800,0.002)};
\end{axis}
\end{tikzpicture}

%% file: bexp.bib
@article{galaz2022wrapped,
  title={Wrapped distributions on homogeneous Riemannian manifolds},
  author={Galaz-Garcia, Fernando and Papamichalis, Marios and Turnbull, Kathryn and Lunagomez, Simon and Airoldi, Edoardo},
  journal={arXiv preprint arXiv:2204.09790},
  year={2022}
}

@article{ganea2018hyperbolic,
  title={Hyperbolic neural networks},
  author={Ganea, Octavian and B{\'e}cigneul, Gary and Hofmann, Thomas},
  journal={Advances in neural information processing systems},
  volume={31},
  year={2018}
}

@article{pennec2006intrinsic,
  title={Intrinsic statistics on Riemannian manifolds: Basic tools for geometric measurements},
  author={Pennec, Xavier},
  journal={Journal of Mathematical Imaging and Vision},
  volume={25},
  number={1},
  pages={127--154},
  year={2006},
  publisher={Springer}
}

@article{rozen2021moser,
  title={Moser flow: Divergence-based generative modeling on manifolds},
  author={Rozen, Noam and Grover, Aditya and Nickel, Maximilian and Lipman, Yaron},
  journal={Advances in neural information processing systems},
  volume={34},
  pages={17669--17680},
  year={2021}
}

@inproceedings{bose2020latent,
  title={Latent variable modelling with hyperbolic normalizing flows},
  author={Bose, Joey and Smofsky, Ariella and Liao, Renjie and Panangaden, Prakash and Hamilton, Will},
  booktitle={International conference on machine learning},
  pages={1045--1055},
  year={2020},
  organization={PMLR}
}

@inproceedings{nagano2019wrapped,
  title={A wrapped normal distribution on hyperbolic space for gradient-based learning},
  author={Nagano, Yoshihiro and Yamaguchi, Shoichiro and Fujita, Yasuhiro and Koyama, Masanori},
  booktitle={International conference on machine learning},
  pages={4693--4702},
  year={2019},
  organization={PMLR}
}

@article{nickel2017poincare,
  title={Poincar{\'e} embeddings for learning hierarchical representations},
  author={Nickel, Maximillian and Kiela, Douwe},
  journal={Advances in neural information processing systems},
  volume={30},
  year={2017}
}

@article{lou2020neural,
  title={Neural manifold ordinary differential equations},
  author={Lou, Aaron and Lim, Derek and Katsman, Isay and Huang, Leo and Jiang, Qingxuan and Lim, Ser Nam and De Sa, Christopher M},
  journal={Advances in Neural Information Processing Systems},
  volume={33},
  pages={17548--17558},
  year={2020}
}

@article{mathieu2020riemannian,
  title={Riemannian continuous normalizing flows},
  author={Mathieu, Emile and Nickel, Maximilian},
  journal={Advances in neural information processing systems},
  volume={33},
  pages={2503--2515},
  year={2020}
}

@article{skopek2019mixed,
  title={Mixed-curvature variational autoencoders},
  author={Skopek, Ondrej and Ganea, Octavian-Eugen and B{\'e}cigneul, Gary},
  journal={arXiv preprint arXiv:1911.08411},
  year={2019}
}

@inproceedings{chen2023flow,
  title={Flow matching on general geometries},
  author={Chen, Ricky TQ and Lipman, Yaron},
  booktitle={International Conference on Learning Representations},
  volume={2024},
  pages={47922--47945},
  year={2024}
}

@article{davidson2018hyperspherical,
  title={Hyperspherical variational auto-encoders},
  author={Davidson, Tim R and Falorsi, Luca and De Cao, Nicola and Kipf, Thomas and Tomczak, Jakub M},
  journal={arXiv preprint arXiv:1804.00891},
  year={2018}
}


%% file: rc_extra.bib
@inproceedings{debortoli2022riemannian,
  title={Riemannian score-based generative modelling},
  author={De Bortoli, Valentin and Mathieu, Emile and Hutchinson, Michael and Thornton, James and Teh, Yee Whye and Doucet, Arnaud},
  booktitle={Advances in Neural Information Processing Systems},
  volume={35},
  year={2022}
}

@inproceedings{mathieu2019poincare,
  title={Continuous hierarchical representations with {P}oincar\'e variational auto-encoders},
  author={Mathieu, Emile and Le Lan, Charline and Maddison, Chris J. and Tomioka, Ryota and Teh, Yee Whye},
  booktitle={Advances in Neural Information Processing Systems},
  volume={32}, year={2019}, note={arXiv:1901.06033}
}

@inproceedings{falorsi2019lie,
  title={Reparameterizing distributions on {L}ie groups},
  author={Falorsi, Luca and de Haan, Pim and Davidson, Tim R. and Forr\'e, Patrick},
  booktitle={Proceedings of the 22nd International Conference on Artificial Intelligence and Statistics},
  year={2019}, note={arXiv:1903.02958}
}

@article{said2017riemannian,
  title={Riemannian Gaussian distributions on the space of symmetric positive definite matrices},
  author={Said, Salem and Bombrun, Lionel and Berthoumieu, Yannick and Manton, Jonathan H},
  journal={IEEE Transactions on Information Theory},
  volume={63},
  number={4},
  pages={2153--2170},
  year={2017},
  publisher={IEEE}
}
